\pdfoutput=1
\documentclass[11pt]{article}

\usepackage[preprint]{acl}

\usepackage{times}
\usepackage{latexsym}
\usepackage{subcaption}
\usepackage{makecell}
\usepackage{listings}
\usepackage{xcolor}
\usepackage{proof}
\usepackage[T1]{fontenc}
\usepackage[utf8]{inputenc}

\usepackage{microtype}

\usepackage{inconsolata}

\usepackage{graphicx}
\usepackage{amsmath}
\usepackage{cleveref}
\usepackage{booktabs}
\usepackage{xcolor}      
\usepackage{booktabs}   
\usepackage{algorithm}
\usepackage{siunitx} 
\usepackage{amssymb}
\usepackage{multirow}
\usepackage{multicol}
\usepackage{algpseudocode}  
\usepackage{amssymb}  
\title{ID-Align: RoPE-Conscious Position Remapping for Dynamic High-Resolution Adaptation in Vision-Language Models}

\author{Bozhou Li\thanks{\,Corresponding author.} \\
Peking University \\
Beijing, China \\
\texttt{libozhou@pku.edu.cn}\\
\And
Wentao Zhang \\
Peking University \\
Beijing, China \\
\texttt{wentao.zhang@pku.edu.cn}
}

\begin{document}
\maketitle
\begin{abstract}
Currently, a prevalent approach for enhancing Vision-Language Models (VLMs) performance is to encode both the high-resolution version and the thumbnail of an image simultaneously. While effective, this method generates a large number of image tokens. When combined with the widely used Rotary Position Embedding (RoPE), its long-term decay property hinders the interaction between high-resolution tokens and thumbnail tokens, as well as between text and image. To address these issues, we propose \textbf{ID-Align}, which alleviates these problems by reordering position IDs. In this method, high-resolution tokens inherit IDs from their corresponding thumbnail token while constraining the overexpansion of positional indices. Our extensive experiments conducted within the LLaVA-Next framework demonstrate that ID-Align delivers comprehensive improvements: it not only achieves notable quantitative gains across multiple benchmarks, highlighted by a significant $6.09\%$ enhancement on MMBench’s relation reasoning tasks, but also qualitatively improves the model's attention distribution, making it more interpretable.
Our code is available at \url{https://github.com/zooblastlbz/ID-Align}.
\end{abstract}
\section{Introduction}

The swift advancement in large language models (LLMs) \citep{achiam2023gpt,cai2024internlm2,yang2024qwen2,liu2024deepseek} has not only revolutionized natural language processing but also catalyzed the emergence of vision-language models (VLMs) \citep{liu2024visual,wu2024deepseek,chen2024far,li2023blip,wang2024qwen2}. In the architecture of these advanced VLMs, visual encoders—such as Vision Transformers (ViTs) \citep{dosovitskiy2020image} employing training objectives like CLIP \citep{radford2021learning} or SigLip \citep{zhai2023sigmoid}—are primarily utilized to encode images. Subsequently, mechanisms such as Multi-Layer Perceptrons (MLPs) \citep{liu2024visual} or Q-Former \citep{li2023blip} are employed to fuse the encoded visual information with textual data. This fused multimodal information is then processed by the LLM, enabling comprehensive understanding and contextually relevant response generation across both visual and textual domains \citep{yin2023survey}.

In the pursuit of developing more effective VLMs, researchers are undertaking multifaceted efforts, including curating higher-quality training datasets \citep{bai2024survey,liu2025synthvlm,chen2026avocado,li2025gran,chen2026diadem} and refining model architectures \citep{cha2024honeybee,li2026unseen}. Beyond these strategies, another approach explored to enhance model performance involves upscaling an input image to a higher resolution before encoding, while concurrently processing a low-resolution version as a thumbnail \citep{dai2024nvlm,deitke2024molmo,wu2024deepseek,chen2024expanding,liu2024improved}. The image tokens derived from both the thumbnail and the high-resolution image are then concatenated and fed into the LLM. This technique is commonly referred to as dynamic high-resolution adaptation.

Despite its straightforwardness and effectiveness, this dynamic high-resolution adaptation method exhibits several critical shortcomings. Encoding high-resolution images inherently generates a large number of image tokens. Consequently, the application of Rotary Position Embedding (RoPE) \citep{su2024roformer}, a prevalent position encoding method, can pose specific challenges due to its characteristic \textbf{long-term decay property}, which posits that attention scores between query and key diminish as their relative distance increases. Although generally assumed to be valid, some researchers have contested this property \citep{barbero2024round}. Our further analysis reveals that, based purely on RoPE's mathematical formulation, its effective behavior (e.g., long-term decay, growth, or more complex patterns) can vary depending on the specific distributions of the query ($\mathbf{q}$) and key ($\mathbf{k}$) vectors. Furthermore, our empirical experiments confirm that, under the actual distributions of $\mathbf{q}$ and $\mathbf{k}$ observed in LLMs, RoPE indeed exhibits this long-term decay property.

This property may lead to:
\begin{itemize}
    \item \textbf{Hinders image-text interaction:} The substantial increase in image embeddings resulting from high-resolution strategies can impede effective interaction between text and image embeddings. This issue is particularly pronounced for image embeddings whose sequential positions are distant from the text embeddings.
    \item \textbf{Loss of Multi-Resolution Correspondence:} A spatial correspondence should exist between high-resolution image tokens and their thumbnail counterparts, where two tokens are defined as corresponding if their encoded regions spatially overlap. However, RoPE's long-term decay property can disrupt this crucial relationship.
\end{itemize}

To address these issues, we propose \textbf{ID-Align}, a novel method that strategically rearranges the position IDs of image tokens. By assigning identical positional IDs to corresponding high-resolution and thumbnail image embeddings, ID-Align preserves their inter-resolution correspondence. This approach not only maintains the crucial relationship between high-resolution and thumbnail tokens but also mitigates the excessive inflation of position ID magnitudes that can arise from the large number of image embeddings in high-resolution strategies. Our experiments, conducted on the LLaVA-Next \citep{liu2024llavanext} architecture, demonstrate that ID-Align significantly enhances model capabilities, particularly concerning fine-grained perception of global information. Our contributions can be summarized into the following three points:
\begin{itemize}
    \item We analyze the mathematical properties of RoPE, demonstrating that its long-term decay property is contingent upon the specific distributions of $\mathbf{q}$ and $\mathbf{k}$ vectors. We further conduct empirical experiments showing that within LLMs, RoPE indeed imparts this long-term decay property to the model's attention mechanism.
    \item We first analyze the adverse effects of the long-term decay property of RoPE when increasing the number of image embeddings using the aforementioned super-resolution methods.
    \item On this basis, we introduce \textbf{ID-Align}, a technique for reorganizing position IDs. This method is aimed at maintaining the correspondence between image embeddings across different resolutions and mitigating the excessive growth of position IDs caused by dynamic adjustments to higher resolutions. Our experiments on the architecture and datasets of LLaVA-Next confirm the effectiveness of ID-Align.
\end{itemize}

\begin{figure*}
    \centering
    \begin{subfigure}[b]{1\textwidth}
    \includegraphics[width=1\linewidth]{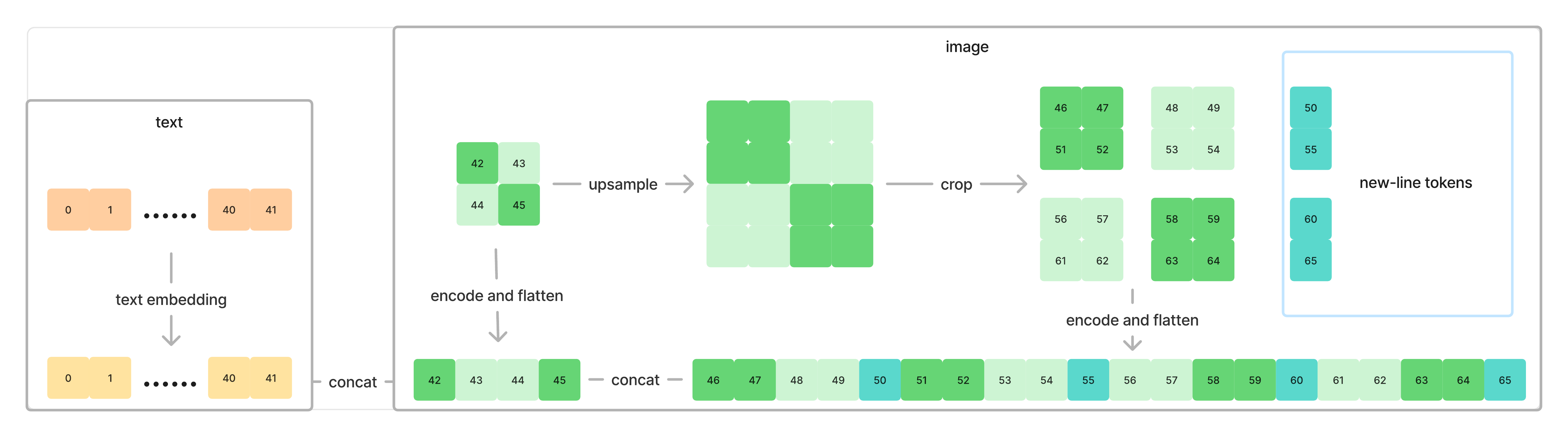}
    \caption{The original method}
    \label{fig:1a}
    \end{subfigure}
    \centering
    \begin{subfigure}[b]{1\textwidth}
    \includegraphics[width=1\linewidth]{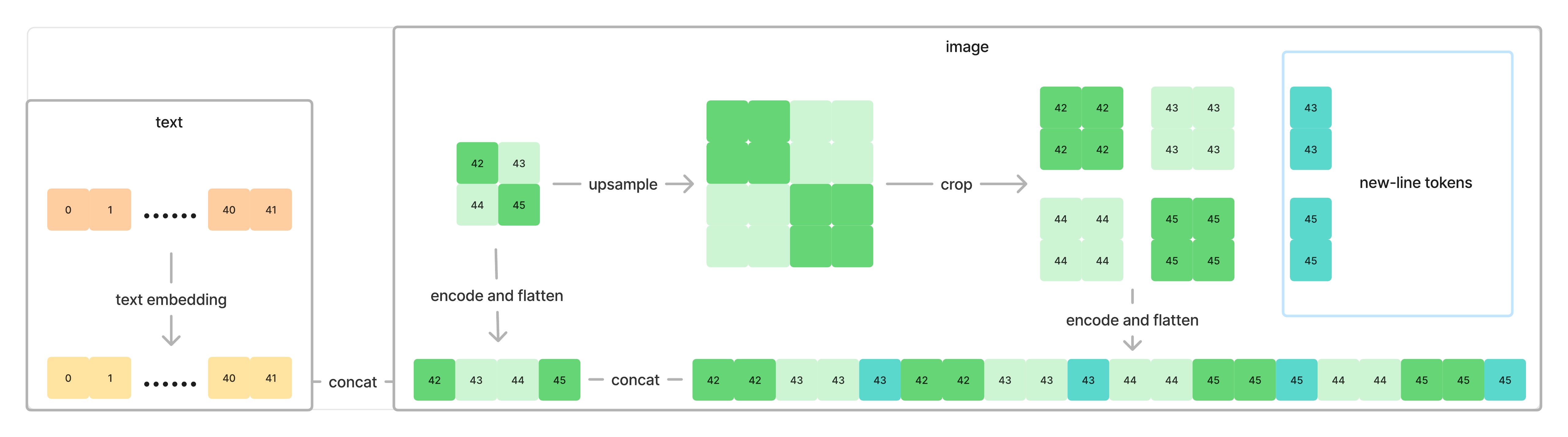}
    \caption{ID-Align}
    \label{fig:1b}
    \end{subfigure}
    \caption{Intuitive presentation of the original high-resolution method and ID-Align.}
        \label{fig:1}
\end{figure*}
\section{Background \& Related Work}

\subsection{Vision Language Model}

 Currently, the mainstream approach to build VLMs is to employ a projector to connect a pre-trained LLM with a visual encoder, thereby enabling the LLM to interpret visual information \citep{zhang2024mm}. For image inputs $I_{image}$, it is usual to first encode them using vision encoders such as SigLIP \citep{zhai2023sigmoid} or CLIP \citep{radford2021learning}  ViT \citep{dosovitskiy2020image}:
\begin{equation}
    F_{image}=VE(I_{image})
\end{equation}
Subsequently, the projector processes the encoded image features $F_{image}$:
\begin{equation}
    P_{image}=Projector(F_{image},I_{text})
\end{equation}
where $I_{text}$ represents the text input. In certain architectures, such as BLIP-2 \citep{li2023blip}, $I_{text}$ also interacts with $F_{image}$ at this stage.
Following this, the LLM backbone processes  $I_{text}$ alongside $P_{image}$, generating the corresponding output:
\begin{equation}
    Output=LLM(I_{text},P_{image})
\end{equation}
The architecture of the projector has many possible designs, and currently, a mainstream choice is to use a two-layer Multilayer Perceptron (MLP) to process $F_{image}$ independently of $I_{text}$, as exemplified by the LLaVA architecture \citep{liu2024visual}:
\begin{equation}
     P_{image}=MLP(F_{image})
\end{equation}
\subsection{Dynamic High-resolution}

\begin{figure}[ht]
    \centering
    \includegraphics[width=1\linewidth]{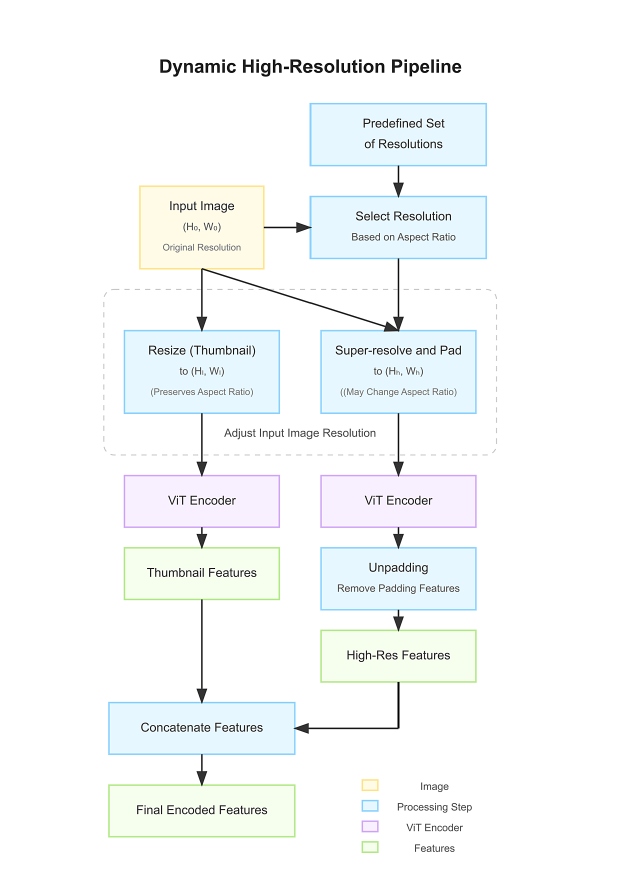}
    \caption{Flowchart of the Dynamic High-Resolution Method}
    \label{fig:high-res}
\end{figure}

While VLMs exhibit remarkable performance across diverse domains, they possess inherent limitations. These are sometimes characterized using the phrase `VLMs are blind' \citep{rahmanzadehgervi2024vision}, denoting their deficiencies in areas such as fine-grained perception and spatial understanding. One effective method is the dynamic high-resolution approach, the process of which is illustrated in Figure \ref{fig:high-res} and includes the following steps:

The current mainstream pipeline is as follows:
\begin{itemize}
    \item Set a Predefined Set of Resolutions. For instance, if the ViT used in a VLM is suitable for processing images of size (336, 336), this set of resolutions could be defined as [(672, 672), (336, 672), (672, 336), (1008, 336), (336, 1008)].
    \item Select Appropriate Resolution. Given an input image with dimensions $(H_0, H_0)$, the most suitable resolution is selected from a set of predefined resolutions based on its aspect ratio.
    \item Adjust Input Image Resolution. For an input image with original resolution $(H_0, W_0)$, two resolution adjustments are applied: first, super-resolving it from its original resolution to a selected higher resolution $(H_h, W_h)$ to obtain a high-resolution image; and second, resizing it to a resolution $(H_l, W_l)$ suitable for the ViT to serve as a thumbnail. The former process often preserves the original image's aspect ratio, filling the remaining regions with blank space, while the latter generally does not.
    \item Encode Image.
ViT is used to encode the high-resolution image and its thumbnail separately. For the encoded features of the high-resolution image, an unpadding stage is typically required to remove the features corresponding to the padding regions. The resulting encoded features are then concatenated to obtain the final encoding.
\end{itemize}

This method is used by various leading VLMs \citep{zhu2025internvl3,liu2024nvila,deitke2024molmo,wu2024deepseek,chen2024expanding,liu2024improved}. When VLMs use a fixed-size ViT for encoding, to handle high-resolution images, the common approach is to divide the high-resolution image into patches or crops, encode each separately, and then rearrange the encoded results. Tokens, such as new-line tokens or separators, are also typically added at appropriate positions. This process can be seen in Figure \ref{fig:1a}.

\subsection{RoPE}
The sequential nature of natural language is pivotal for understanding its semantics. However, the attention mechanism employed in the Transformer \citep{vaswani2017attention} architecture does not inherently capture this sequential information. Consequently, it is essential to incorporate positional encoding within the Transformer model to enable the processing of sequence-dependent information. For the query 
$\boldsymbol{q} $ with the position ID $m$ and key $\boldsymbol{k}$ with  the position ID $n$, positional encoding is applied to incorporate positional information into them:
\begin{equation}
    \hat{\boldsymbol{q}}=PE(\boldsymbol{q},m)  ,\hat{\boldsymbol{k}}=PE(\boldsymbol{k},n) 
\end{equation}

Positional encoding can be implemented in various ways \citep{gehring2017convolutional,liu2020learning,shaw2018self,dai2019transformer,raffel2020exploring,he2020deberta,wang2019encoding}. Nowadays, in the choice of positional encoding methods, Rotary Position Embedding (RoPE) \citep{su2024roformer} has become a prevalent encoding method. The implementation of RoPE is as follows:

\begin{equation}
    RoPE(\textbf{q},m)= \mathcal{R}_m\textbf{q}
\end{equation}
where:
\begin{equation}
    \mathcal{R}_m = 
    \begin{pmatrix}
    A_0 & 0 & \cdots & 0 \\
    0 & A_1 & \cdots & 0 \\
    \vdots & \vdots & \ddots & \vdots \\
    0 & 0 & \cdots & A_{d/2-1}
    \end{pmatrix}
\end{equation}
\begin{equation}
     A_i = 
    \begin{pmatrix}
    \cos m\theta_i & -\sin m\theta_i \\
    \sin m\theta_i & \cos m\theta_i
\end{pmatrix}   
\end{equation}
\begin{equation}
    \theta_i= \theta^{-\frac{2i}{d}}
\end{equation}
Where $d$ is the dimensionality of $\textbf{q}$, $\theta$  is a hyperparameter, typically taking values ranging from $10^4 $ to $10^7$.

RoPE exhibits several key characteristics:
\begin{itemize}
    \item RoPE can be described as a form of absolute positional encoding because it uses the absolute positions of tokens during the encoding process. However, it also exhibits properties of relative positional encoding due to its mathematical property:

\begin{equation}
\begin{split}
(\mathcal{R}_m\textbf{q})^T(\mathcal{R}_n\textbf{k}) & = \textbf{q}^T\mathcal{R}_m^T\mathcal{R}_n\textbf{k} \\
& = \textbf{q}^T\mathcal{R}_{n-m}\textbf{k}
\end{split}
\end{equation}

    \item RoPE exhibits a characteristic of long-range decay: for a query $\textbf{q}$ at position $m$ and a key $\textbf{k}$ at position $n$, after encoding with RoPE, the dot product $(\mathcal{R}_m\textbf{q})^T(\mathcal{R}_n\textbf{k}) $ generally decreases as the absolute value of $|m-n|$ increases. However, this property of RoPE is partially controversial, which we will discuss further in Section \ref{sec:rope}.
    \item The value of $\theta$ controls the positional encoding's sensitivity to positional differences. A smaller $\theta$ makes the model more sensitive to position changes, whereas a larger one facilitates the capture of long-range dependencies. Generally, the value of $\theta$ should increase as the training length increases \citep{men2024base}.
\end{itemize}

In the domain of VLMs, researchers are exploring modifications to RoPE to better accommodate multimodal features. Approaches such as CCA \citep{xing2025mitigating} and PyPE \citep{chen2025advancing} aim to reconfigure position IDs from distinct angles, whereas V2PE \citep{Ge2024V2PEIM} narrows the incremental scale of positional encodings specifically for image embeddings. Despite these advancements, none of these proposed methods sufficiently consider the prevalent application of super-resolution techniques—a critical aspect of the current technological landscape.

\section{Analysis}

\subsection{On the long-range decay property of RoPE} \label{sec:rope}
In the RoPE paper \citep{su2024roformer}, the authors theoretically analyzed the long-range decay properties of RoPE:
\begin{align} 
& \left| \sum_{i=0}^{d/2-1} \mathbf{q}_{[2i:2i+1]} \mathbf{k}_{[2i:2i+1]} e^{i(m-n)\theta_i} \right| \nonumber \\
& \leq \left( \max_i |h_{i+1} - h_i| \right) \sum_{i=0}^{d/2-1} |S_{i+1}| 
\label{eq:11}
\end{align}
where:
\begin{align}
    &h_i = \mathbf{q}_{[2i:2i+1]}\mathbf{k}_{[2i:2i+1]}\\
&S_j = \sum_{i=0}^{j-1} e^{i(m-n)\theta_i}
\end{align}
Since the value of $\frac{1}{d/2} \sum_{i=1}^{d/2} |S_i|$
 is decreasing, the above formula indicates that the upper bound of 
$\textbf{q}^T\mathcal{R}_{n-m}\textbf{k}$
 is decreasing as the relative distance $|m-n|$ increases.

They also plotted the  $\textbf{q}^T\mathcal{R}_{n-m}\textbf{k}$
 as a function of their relative distance, specifically for the case where $\textbf{q}$
 and $\textbf{k}$ are all-one vectors, to illustrate RoPE's long-range decay properties.

Although the long-range decay property of RoPE is generally accepted, unlike positional encodings such as ALiBi\citep{press2021train} that explicitly incorporate terms for long-range decay, some researchers have raised questions about this property, and the above inequality is not tight. Some researchers argue that if $\mathbf{q}$ and $\mathbf{k}$ are sampled from a standard multivariate normal distribution, the following formula holds:
\begin{equation}
\mathbb{E}_{\mathbf{q},\mathbf{k}\sim\mathcal{N}(\mathbf{0},\mathbf{I})} [\mathbf{q}^\top \mathcal{R}_{m} \mathbf{k}] = 0 \quad \forall m \in \mathbb{Z}
\end{equation}
leading them to conclude that RoPE does not possess the long-range decay property \citep{barbero2024round}.

 However, their conclusions are based only on their rigorous assumptions. We point out that if $\mathbf{q} \sim \mathcal{N}(\mathbf{\mu_q},\mathbf{I}),\mathbf{k} \sim \mathcal{N}(\mathbf{\mu_{k},\mathbf{I}})$, the following formula holds:
 \begin{equation}
    \mathbb{E}[\mathbf{q}^\top \mathcal{R}_{m} \mathbf{k}] = \mathbf{\mu_q}^T\mathcal{R}_m\mathbf{\mu_k} \quad \forall m \in \mathbb{Z}  \label{eq:15}
 \end{equation}
Furthermore, the trend of $\mathbb{E}[\mathbf{q}^\top \mathcal{R}_{m} \mathbf{k}]$
 with respect to $m$ 
 is dependent on the value of 
$\mathbf{\mu_q},\mathbf{\mu_k}$
, and can be overall increasing or decreasing as $m$
 increases. More detailed results can be found in Appendix~\ref{sec:rope-decay-appendix}.

 Therefore, under the assumption of a normal distribution, we cannot prove that RoPE exhibits the property of long-range decay. However, deep neural networks possess a large number of parameters and are highly complex. During the training process, RoPE also influences model parameter updates, consequently affecting the activation values of query-key pairs. Thus, the simple assumption of a normal distribution is likely not representative of the actual situation.

To investigate whether RoPE exhibits a long-range decay property, we adopted an empirical approach. Specifically, we randomly sampled several data sequences from the WikiText \citep{merity2016pointer} dataset. Then, for each layer, we randomly selected several q-k pairs before applying RoPE. By fixing these token pairs and progressively increasing their relative positions starting from 0, we measured the average inner product at each relative position. The results are shown in the Figure \ref{fig:RoPE}.

 \subsection{Problems with Previous Positional ID Arrangements}
Having empirically demonstrated that RoPE indeed exhibits the long-range decay property in LLMs, we further analyze the issues inherent in previous positional encoding arrangements.
\subsubsection{Disrupt the correspondence between thumbnail and high-resolution images.}
Dynamic high-resolution methods employed by models such as LLaVA-Next simultaneously provide the LLM backbone with both high-resolution images and thumbnails. The high-resolution images furnish the model with fine-grained visual details, while the thumbnails offer global context. Similar to the introduction of RoPE in transformers for NLP to encourage attention mechanisms to focus on nearby tokens, images also exhibit local self-correlation. Consequently, during the interaction between high-resolution image tokens and thumbnail tokens, we aim for the high-resolution image tokens to attend more strongly to their corresponding thumbnail tokens. Here, two tokens are defined as corresponding if the image region encoded by the high-resolution token intersects with the image region encoded by the thumbnail token

However, the specific arrangement of position IDs in this dynamic high-resolution method, coupled with the long-term decay characteristic of RoPE, undermines this corresponding relationship. As shown in Figure \ref{fig:1a},
\begin{itemize}
    \item For a token in the bottom-right corner of the high-resolution image, other tokens within the high-resolution region are relatively closer compared to their corresponding thumbnail tokens. 
    \item For the token in the top-left corner of the high-resolution image, compared to its corresponding thumbnail token, its relative distance to the token in the bottom-right corner of the thumbnail is shorter.
\end{itemize}

As shown in Figure \ref{fig:4b}, when computing the attention distribution from the red region of the high-resolution image towards the thumbnail, the attention can only focus on relevant information in shallow layers, while in deeper layers, attention is concentrated on unrelated areas.

\subsubsection{Disrupts the interaction between text and image}
Dynamic high-resolution methods produce a large number of image tokens. If a conventional position ID arrangement is used, this can result in excessive variation among the position IDs of image tokens corresponding to the same image. Assume a square image is input. In the dynamic high-resolution method, its width and height are scaled up to twice the original dimensions. Compared to approaches that do not use dynamic high resolution, the number of tokens increases by a factor of five, and consequently, the difference in positional encoding among image tokens also expands fivefold.

Effective acquisition of visual information during interaction with user instructions requires engaging with every image token. However, the distance between the top-left image token and the user instruction tokens is significant, causing the user instruction to attend more to the bottom-left corner of the image. The dynamic high-resolution method exacerbates this problem by increasing the difference in position IDs between the top-left and bottom-right tokens.

Furthermore, studies have shown that in VLMs, image tokens inherently receive less attention \citep{chen2024image}. Coupled with RoPE's long-range decay characteristic, the excessive relative position between the top-left token and the user instruction tokens may lead to this part of the information being overlooked or neglected.

\begin{figure*}[!ht]
    \centering
    \begin{subfigure}[b]{1\linewidth}
    \includegraphics[width=1\linewidth]{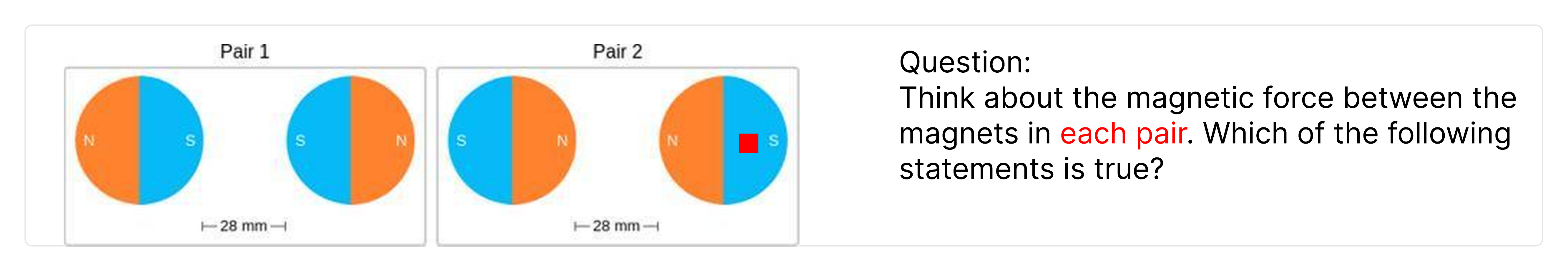}
    \caption{Data example from MMBench, where the red square and red text are used for attention computation.}
    \label{fig:4a}
    \end{subfigure}
    \centering
    \begin{subfigure}[b]{1\textwidth}
    \includegraphics[width=1\linewidth]{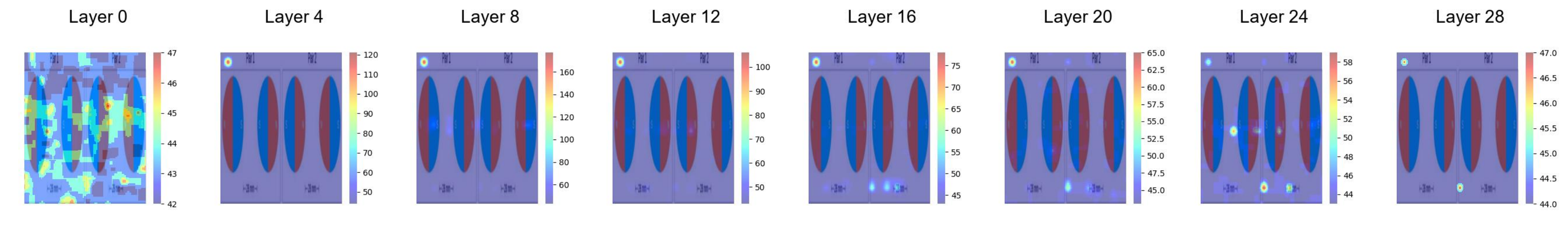}
    \caption{Attention distribution for the red region (w/o ID-Align).}
    \label{fig:4b}
    \end{subfigure}
    \centering
    \begin{subfigure}[b]{1\textwidth}
    \includegraphics[width=1\linewidth]{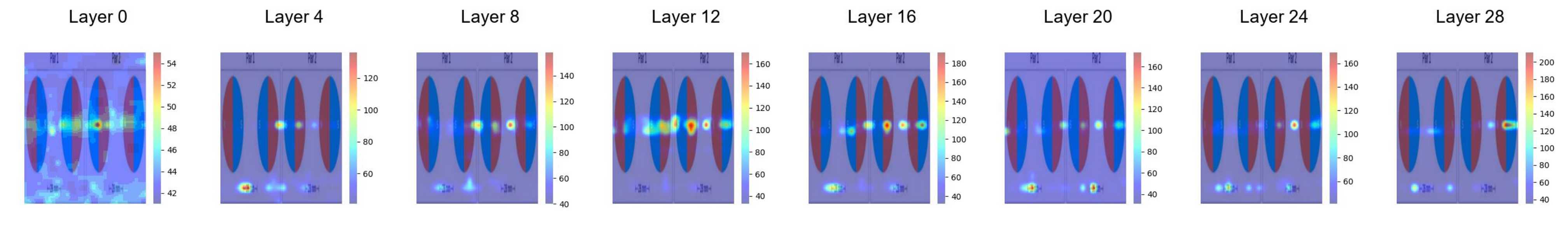}
    \caption{Attention distribution for the red region (w ID-Align).}
    \label{fig:4c}
    \end{subfigure}
    \centering
    \begin{subfigure}[b]{1\textwidth}
    \includegraphics[width=1\linewidth]{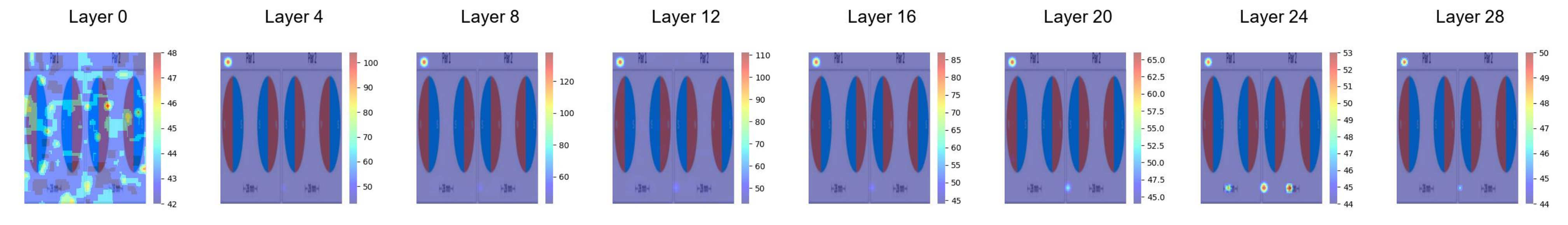}
    \caption{Attention distribution for the red text (w/o ID-Align).}
    \label{fig:4d}
    \end{subfigure}
    \centering
    \begin{subfigure}[b]{1\textwidth}
    \includegraphics[width=1\linewidth]{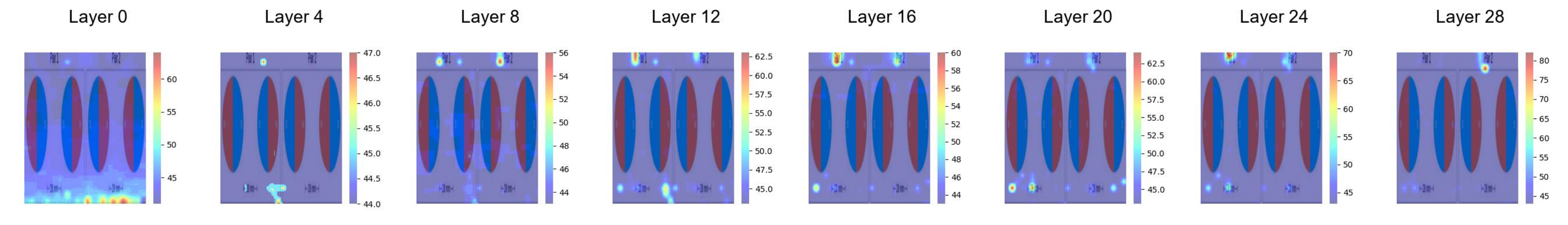}
    \caption{Attention distribution for the red text (w ID-Align).}
    \label{fig:4e}
    \end{subfigure}
    \caption{Attention distributions from the red region in the high-resolution image and the red text towards thumbnail tokens. Figure \ref{fig:4a} shows the data example. Figures \ref{fig:4b} and \ref{fig:4c} depict the attention distribution from the red region, and figures  \ref{fig:4d} and  \ref{fig:4e} show the attention distribution from the red text.}
    \label{fig:4}
\end{figure*}

As shown in Figure \ref{fig:4d}, when computing the attention distribution from `each pair' towards the thumbnail, the attention is neither able to focus on the corresponding text in the image nor on the corresponding object.

\section{Methods}

According to the calculation formula of RoPE, it can be observed that during inference, the relative distance between $\mathbf{q}$ and $\mathbf{k}$ is influenced not by their actual distance in the sequence, but by the difference in their position IDs. Simultaneously, as shown in Section \ref{sec:rope}, increasing the difference between the position IDs of $\mathbf{q}$ and $\mathbf{k}$ can enhance their attention coefficient, while decreasing it can reduce it. Therefore, we propose to alleviate the aforementioned issues by rearranging the position IDs.  Our approach is as follows:

    \begin{itemize}
        \item For the tokens of thumbnails, we adopt the same position IDs as those used in the previously established approach.
        \item For the tokens of high-resolution images, we assign them the same position ID as their corresponding thumbnail image tokens.
    \end{itemize}
    
The difference between our method and the original approach can be seen in Figure \ref{fig:1}. More details are available in Appendix \ref{sec:method details}.

\section{Experiments and Results}

\subsection{Experiments Setup}
We adopted the LLaVA-Next architecture \citep{liu2024llavanext}.
We used the  Vicuna-1.5 7B \citep{zheng2023judging} as the LLM backbone and CLIP ViT-L/14 (336) \citep{radford2021learning} as the vision encoder. Alternatively, we used Qwen-2.5-7B-Instruct \citep{yang2024qwen2} as the backbone and SigLip 400M \citep{zhai2023sigmoid} as the encoder. It is worth noting that the RoPE $\theta$ for the Qwen series models is $10^7$, which is significantly larger than that of the Vicuna model ($10^4$). This indicates that Qwen models are relatively less sensitive to changes in positional IDs. For all evaluations, we consistently employed a greedy decoding strategy. Complete training details and the full set of benchmark results are provided in the Appendix \ref{sec:experiments set up}.

\subsection{Results and Analysis}
\begin{table*}[!ht]
    \centering
    % Define enough columns for the section with more benchmarks (6 benchmarks + Avg = 7 pairs)
    % l (Model) + 6 * (r l) (MMBench, MMStar, RealWorldQA, SEEDB2-Plus, POPE, Avg)
    \begin{tabular}{l r@{\,}l  r@{\,}l r@{\,}l  r@{\,}l  r@{\,}l }
        \toprule
       \textbf{Model}   & \multicolumn{2}{c}{$\textbf{MMBench}_{dev}$} & \multicolumn{2}{c}{\textbf{MMStar}} & \multicolumn{2}{c}{\textbf{RealWorldQA}} & \multicolumn{2}{c}{\textbf{SEEDB2-Plus}} & \multicolumn{2}{c}{\textbf{POPE}@\textbf{ACC}}  \\
        \midrule
        Vicuna \\
        w/o ID-Align &  66.58   & & 36.61 & & 58.43 && 51.38 && 87.97 &  \\
        w/ \: ID-Align & \textbf{68.21} & \scriptsize{(+1.63)}  & \textbf{38.32} & \scriptsize{(+1.71)} & \textbf{59.18} & \scriptsize{(+0.75)} & \textbf{51.56} & \scriptsize{(+0.18)} & \textbf{88.66} & \scriptsize{(+0.69)}  \\
        \midrule
        Qwen \\
        w/o ID-Align & 78.14 & & \textbf{50.53} & & \textbf{64.18} & & 61.00 & & \textbf{89.17} & \\
        w/ \: ID-Align & \textbf{78.48}  & \scriptsize{(+0.34)} & 50.14 &\scriptsize{(-0.39)} & 63.79 & \scriptsize{(-0.39)}& \textbf{62.06} & \scriptsize{(+1.06)}& 89.16 &\scriptsize{(-0.01)} \\
        \midrule
        % Headers for the second set of benchmarks (4 benchmarks + Avg = 5 pairs).
        % The 6th pair of columns from the tabular definition will be left empty.
& \multicolumn{2}{c}{$\mathbf{MME}$} 
& \multicolumn{2}{c}{$\mathbf{AI2D}$}
& \multicolumn{2}{c}{$\mathbf{VQAV2}_{\text{val}}$} 
& \multicolumn{2}{c}{$\mathbf{SQA}_{\text{img}}$} 
& \multicolumn{2}{c}{$\mathbf{Avg}$} \\
         \midrule
         Vicuna \\
         w/o ID-Align &  65.22 & &  65.74 && 79.75  && 69.41 & & 64.57&  \\
         w/ \: ID-Align &  \textbf{65.50} & \scriptsize{(+0.28)} & \textbf{66.39}  & \scriptsize{(+0.65)}& \textbf{80.02} & \scriptsize{(+0.27)}& \textbf{70.70} &\scriptsize{(+1.29)} & \textbf{65.39} & \scriptsize{(+0.82)}  \\
         \midrule
         Qwen \\
         w/o ID-Align & 67.11 && 74.84 && 79.88 && 80.61 & & 71.72 &  \\
         w/ \: ID-Align & \textbf{68.22} & \scriptsize{(+1.11)}& \textbf{75.13} &\scriptsize{(+0.29)}&  \textbf{80.25}  & \scriptsize{(+0.37)}& \textbf{81.06}&\scriptsize{(+0.45)} & \textbf{72.03} & \scriptsize{(+0.31)}  \\
        \bottomrule
    \end{tabular}
        \caption{Performance on Different Benchmarks with and without ID-Align}
        \label{tab:main-results}
\end{table*}

% 75.56
\begin{table*}[!ht]
    \centering
    \begin{tabular}{l c@{\,} l c@{\,} l c@{\,} l c@{\,} l c@{\,} l c@{\,} l}
        \toprule
        \textbf{Model} 
         & \textbf{CP} & & \textbf{FP-S} & & \textbf{FP-C} & &\textbf{AR} & &\textbf{RR} & &\textbf{LR} \\
        \midrule
        Vicuna & & & & & & \\
        w/o ID-Align & 79.39 & & 70.31 & & 58.04 & & 69.35 & & 60.87 & & \textbf{36.44} & \\
        w/ \: ID-Align &\textbf{81.76} & \scriptsize{(+2.37)} & \textbf{71.67} & \scriptsize{(+1.36)} & \textbf{59.44} & \scriptsize{(+1.40)} & \textbf{69.85} & \scriptsize{(+0.50)} & \textbf{66.96} & \scriptsize{(+6.09)} & \ 34.75 & \scriptsize{(-1.69)} \\

        \midrule
        Qwen & & & & & & \\
        w/o ID-Align & \textbf{83.73} && \textbf{81.91} && 71.26 && \textbf{84.38} && 75.83 && 56.65 & \\
        w/ \: ID-Align & 82.87 &\scriptsize{(-0.86)}& \textbf{81.91} &\scriptsize{(+0.00)}& \textbf{72.87} &\scriptsize{(+1.61)}& 83.33 &\scriptsize{(-1.05)}& \textbf{77.72} &\scriptsize{(+1.89)}& \textbf{59.54} & \scriptsize{(+2.89)}\\
        \bottomrule
    \end{tabular}
        \caption{The table presents the results on sub-metrics from the MMBench-Dev. Specifically, \textbf{CP} stands for Coarse Perception, \textbf{FP-C} represents Fine-grained Perception (cross-instance), \textbf{FP-S} denotes Fine-grained Perception (single-instance), \textbf{AR} refers to Attribute Reasoning, \textbf{LR} indicates Logical Reasoning, \textbf{RR} represents Relation Reasoning. }
        \label{tab:performance_comparison}
\end{table*}
To intuitively demonstrate the optimization effect of ID-Align on attention distribution, we visualized attention heatmaps. As shown in Figure \ref{fig:4}, after applying ID-Align, the model's attention becomes significantly more focused on regions relevant to the queries, rather than being scattered across unrelated areas. Specifically, comparing Figure \ref{fig:4c} with \ref{fig:4b}, when the query is the red region in the image, the attention is no longer confined to irrelevant areas but precisely focuses on the magnets in the image. Similarly, comparing Figure \ref{fig:4e} with \ref{fig:4d}, when the query is the text `each pair', the attention accurately concentrates on the corresponding text portion in the thumbnail.

The primary experimental results are shown in Table ~\ref{tab:main-results}. As can be observed from the table, the adoption of ID-Align has led to improvements in the model's performance metrics across various benchmarks. When using Vicuna and CLIP as pre-training models, there was a notable improvement across all benchmarks. These benchmarks cover a broad spectrum of capabilities, indicating the effectiveness of our approach. When employing Qwen2.5, which has a RoPE $\theta$ value of $10^7$, and SigLIP as the base models, the performance gains were observed to decrease, and there was a decline in performance on several benchmarks. This observation aligns with our analysis, which suggests that these models are relatively insensitive to changes in positional encoding. However, after adopting ID-Align, the overall performance of the model showed an increasing trend. 

To further investigate which specific capabilities contributed most to the observed growth in benchmark performance, we have detailed the changes in various sub-metrics of MMBench, as shown in Table \ref{tab:performance_comparison}. 
We have also listed the subtasks of MMBench in Appendix \ref{sec:mmbench}. As can be observed, when using Vicuna as the LLM backbone, although all sub-indicators showed improvement, the most significant growth was seen in the RR metrics. Meanwhile, when employing Qwen as the LLM backbone, it was the FP-C, RR, and LR metrics that maintained their growth. These metrics are all related to global information.

%We also evaluated the performance of ID-Align in a training-free scenario on MMBench, where ID-Align was not employed during the training phase but was applied during inference. The results are shown in Table \ref{tab:performance_comparison}. It can be observed from the table that, in the training-free setting, the model's capability for fine-grained cross-instance perception and reasoning still improves, albeit with a smaller margin. However, its performance regarding single-instance tasks declines.

\section{Conclusion}
In this paper, we analyze the potential issues of the dynamic high-resolution strategies adopted by current VLMs. Based on our analysis, we propose \textbf{ID-Align}: a method that aligns the position IDs of high-resolution embeddings with their corresponding low-resolution embeddings, preserving their relationship and constraining excessive growth in position IDs. We conducted experiments to demonstrate the effectiveness of our approach.
\section{Limitations}

Limitations of our work include: we did not investigate the performance of our method when combined with token compression techniques. We also did not examine the performance of our method when integrated with ViT that inherently support dynamic resolution.

\bibliography{custom}

\appendix
\section{Long-term decay of RoPE}
\label{sec:rope-decay-appendix}
The proof of Equation (\ref{eq:11}) is as follows:\\
Let:
\begin{equation}
\begin{split}
   & h_i = \mathbf{q}_{[2i:2i+1]} \mathbf{k}_{[2i:2i+1]} \\
   & S_j = \sum_{i=0}^{j-1} e^{i(m-n)\theta_i}
\end{split}
\end{equation}
Setting $h_{d/2} = 0$ and $S_0 = 0$, with the Abel transformation, we have:
\begin{equation} 
\begin{split} % Using split, with & at the beginning of each line
& \sum_{i=0}^{d/2-1} \mathbf{q}_{[2i:2i+1]} \mathbf{k}_{[2i:2i+1]}^* e^{i(m-n)\theta_i} \\
& = \sum_{i=0}^{d/2-1} h_i (S_{i+1} - S_i) \\
& = - \sum_{i=0}^{d/2-1} S_{i+1} (h_{i+1} - h_i).
\end{split}
\end{equation}
Thus,
\begin{align} % Using align, with & at the beginning of each line
& \left| \sum_{i=0}^{d/2-1} \mathbf{q}_{[2i:2i+1]} \mathbf{k}_{[2i:2i+1]}^* e^{i(m-n)\theta_i} \right| \nonumber \\
& = \left| \sum_{i=0}^{d/2-1} S_{i+1} (h_{i+1} - h_i) \right| \nonumber \\
& \leq \sum_{i=0}^{d/2-1} |S_{i+1}| |(h_{i+1} - h_i)| \nonumber \\
& \leq \left( \max_i |h_{i+1} - h_i| \right) \sum_{i=0}^{d/2-1} |S_{i+1}| 
\end{align}

The proof of Equation (\ref{eq:15}) is as follows: \\

Let $\mathbf{q} \sim \mathcal{N}(\mathbf{\mu}_q,\mathbf{I})$ and $\mathbf{k} \sim \mathcal{N}(\mathbf{\mu}_k,\mathbf{I})$ be independent random vectors 

We use the law of total expectation, conditioning on 
$\mathbf{k}$:

\begin{align*}
\mathbb{E}[\mathbf{q}^\top \mathcal{R}_{m} \mathbf{k}] &= \mathbb{E}_\mathbf{k} \left[ \mathbb{E}_{\mathbf{q}|\mathbf{k}}[\mathbf{q}^\top \mathcal{R}_{m} \mathbf{k} \mid \mathbf{k}] \right] \\
&= \mathbb{E}_\mathbf{k} \left[ \mathbb{E}[\mathbf{q}^\top \mid \mathbf{k}] \mathcal{R}_{m} \mathbf{k} \right]\ \\
&= \mathbb{E}_\mathbf{k} \left[ \mathbb{E}[\mathbf{q}^\top] \mathcal{R}_{m} \mathbf{k} \right] \\
&= \mathbb{E}_\mathbf{k} \left[ \boldsymbol{\mu}_q^\top \mathcal{R}_{m} \mathbf{k} \right]\\
&= \boldsymbol{\mu}_q^\top \mathcal{R}_{m} \mathbb{E}_\mathbf{k}[\mathbf{k}] \\
&= \boldsymbol{\mu}_q^\top \mathcal{R}_{m} \boldsymbol{\mu}_k. 
\end{align*}

The $i$-th $2 \times 2$ block of $\mathcal{R}_m$, denoted $\mathcal{R}_m^{(i)}$, is given by:
\[
\mathcal{R}_m^{(i)} = \begin{pmatrix} \cos(m\theta_i) & -\sin(m\theta_i) \\ \sin(m\theta_i) & \cos(m\theta_i) \end{pmatrix}
\]

First, the product $\mathcal{R}_m \boldsymbol{\mu}_k$ results in a vector where the components corresponding to the $i$-th 2D block are:
\begin{align*}
(\mathcal{R}_m \boldsymbol{\mu}_k)_{2i-1} &= \mu_{k,2i-1} \cos(m\theta_i) \\
&\quad - \mu_{k,2i} \sin(m\theta_i) \\
(\mathcal{R}_m \boldsymbol{\mu}_k)_{2i} &= \mu_{k,2i-1} \sin(m\theta_i) \\
&\quad + \mu_{k,2i} \cos(m\theta_i)
\end{align*}
The dot product $\boldsymbol{\mu}_q^\top (\mathcal{R}_m \boldsymbol{\mu}_k)$ is then:
\[
\boldsymbol{\mu}_q^\top \mathcal{R}_m \boldsymbol{\mu}_k = \sum_{j=1}^d \mu_{q,j} (\mathcal{R}_m \boldsymbol{\mu}_k)_j
\]
Grouping the summation by the $d/2$ two-dimensional blocks:
\begin{align*}
&\boldsymbol{\mu}_q^\top \mathcal{R}_m \boldsymbol{\mu}_k \\
&= \sum_{i=1}^{d/2} \big[ \mu_{q,2i-1} (\mathcal{R}_m \boldsymbol{\mu}_k)_{2i-1} \\
&\qquad + \mu_{q,2i} (\mathcal{R}_m \boldsymbol{\mu}_k)_{2i} \big] \\
&= \sum_{i=1}^{d/2} \big[ \mu_{q,2i-1} (\mu_{k,2i-1} \cos(m\theta_i) \\
&\qquad\quad - \mu_{k,2i} \sin(m\theta_i)) \\
&\qquad + \mu_{q,2i} (\mu_{k,2i-1} \sin(m\theta_i) \\
&\qquad\quad + \mu_{k,2i} \cos(m\theta_i)) \big]
\end{align*}

Rearranging the terms within the parentheses and grouping by $cos(m\theta_i)$ and $sin(m\theta_i)$:
\begin{align*}
&\boldsymbol{\mu}_q^\top \mathcal{R}_{m} \boldsymbol{\mu}_k =\\
& \sum_{i=1}^{d/2} \Big[ (\mu_{q,2i-1}\mu_{k,2i-1} + \mu_{q,2i}\mu_{k,2i})\cos(m\theta_i) \\
&\quad + (\mu_{q,2i}\mu_{k,2i-1} - \mu_{q,2i-1}\mu_{k,2i})\sin(m\theta_i) \Big]
\end{align*}

To simplify notation, let:
\begin{align*}
A_i &= \mu_{q,2i-1} \mu_{k,2i-1} + \mu_{q,2i} \mu_{k,2i} \\
B_i &= \mu_{q,2i} \mu_{k,2i-1} - \mu_{q,2i-1} \mu_{k,2i}
\end{align*}
The expression then becomes:
\[
\boldsymbol{\mu}_q^\top \mathcal{R}_m \boldsymbol{\mu}_k = \sum_{i=1}^{d/2} (A_i \cos(m\theta_i) + B_i \sin(m\theta_i))
\]
From this expression, we cannot derive the trend of $\boldsymbol{\mu}_q^\top \mathcal{R}_m \boldsymbol{\mu}_k$ as $m$ changes. Next, we will demonstrate experimentally that $\boldsymbol{\mu}_q^\top \mathcal{R}_m \boldsymbol{\mu}_k$ exhibits different trends with respect to $m$ depending on the values of $\boldsymbol{\mu}_q$ and $\boldsymbol{\mu}_k$.

\begin{figure*}[!htbp]
    \centering
    \includegraphics[width=1\linewidth]{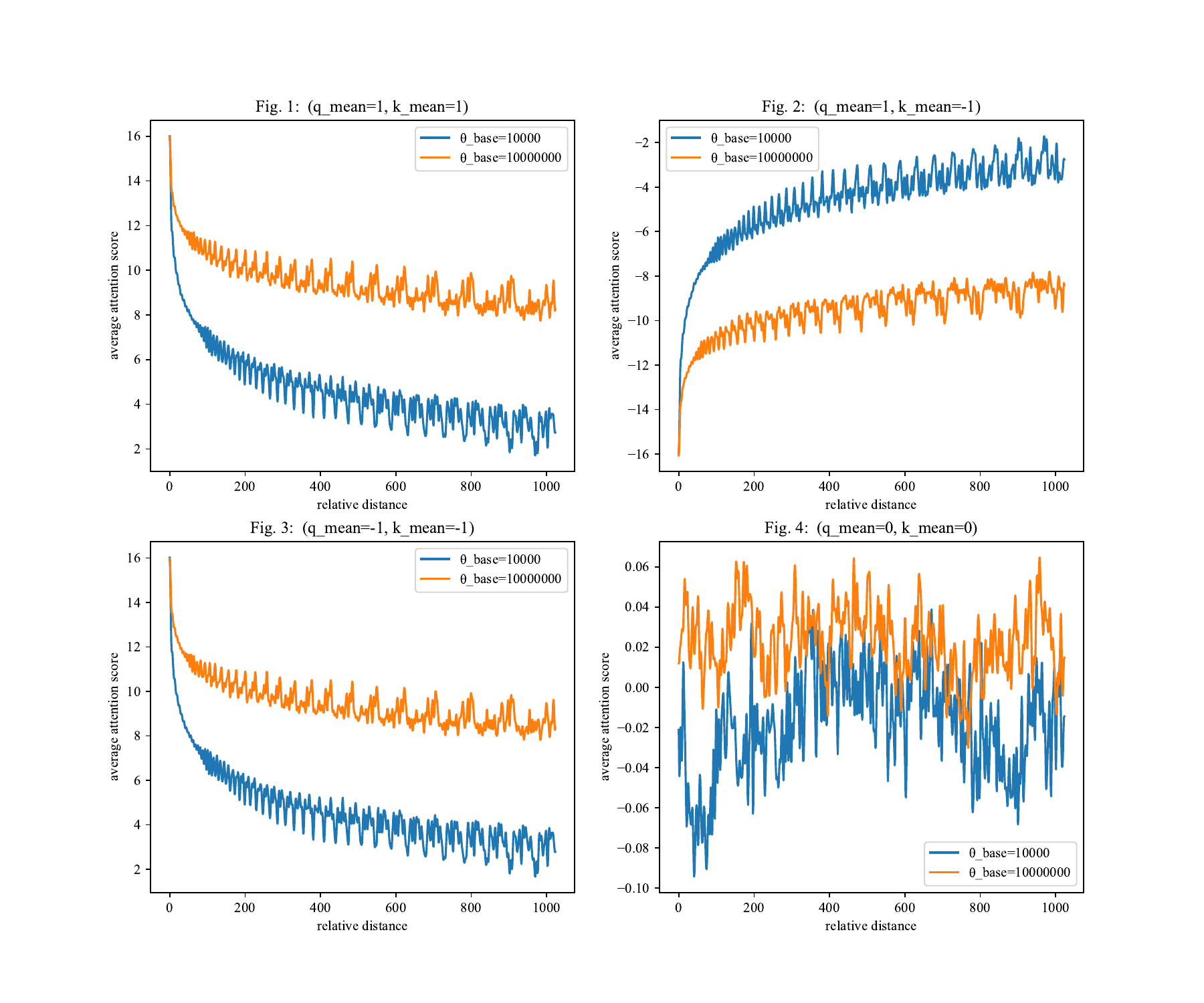}
    \caption{Simulation of RoPE's Long-term Properties under Different $\mathbf{\mu}_q$ and $\mathbf{\mu}_k$}
    \label{fig:rope_theta}
\end{figure*}

For each component of $\mathbf{q}$and  $\mathbf{k}$, we sampled from normal distributions with the same mean and a standard deviation of 1. Different mean values were set for  $\mathbf{q}$ and  $\mathbf{k}$ in each experimental run. Then, we set the relative distance between them to different values and calculated their attention scores. For each choice of mean value, we simulated 1000 times and averaged the results at each relative position. We experimented with two values of $\theta,10^4$ and $10^7$. The results are shown in Figure \ref{fig:rope_theta}. The experiments reveal that different values of $\mathbf{\mu}_q$ and $\mathbf{\mu}_k$ influence the long-term properties of RoPE, and a small value of $\theta$ increases the positional sensitivity of dot-product attention.

 \begin{figure*}
     \centering
     \includegraphics[width=\linewidth]{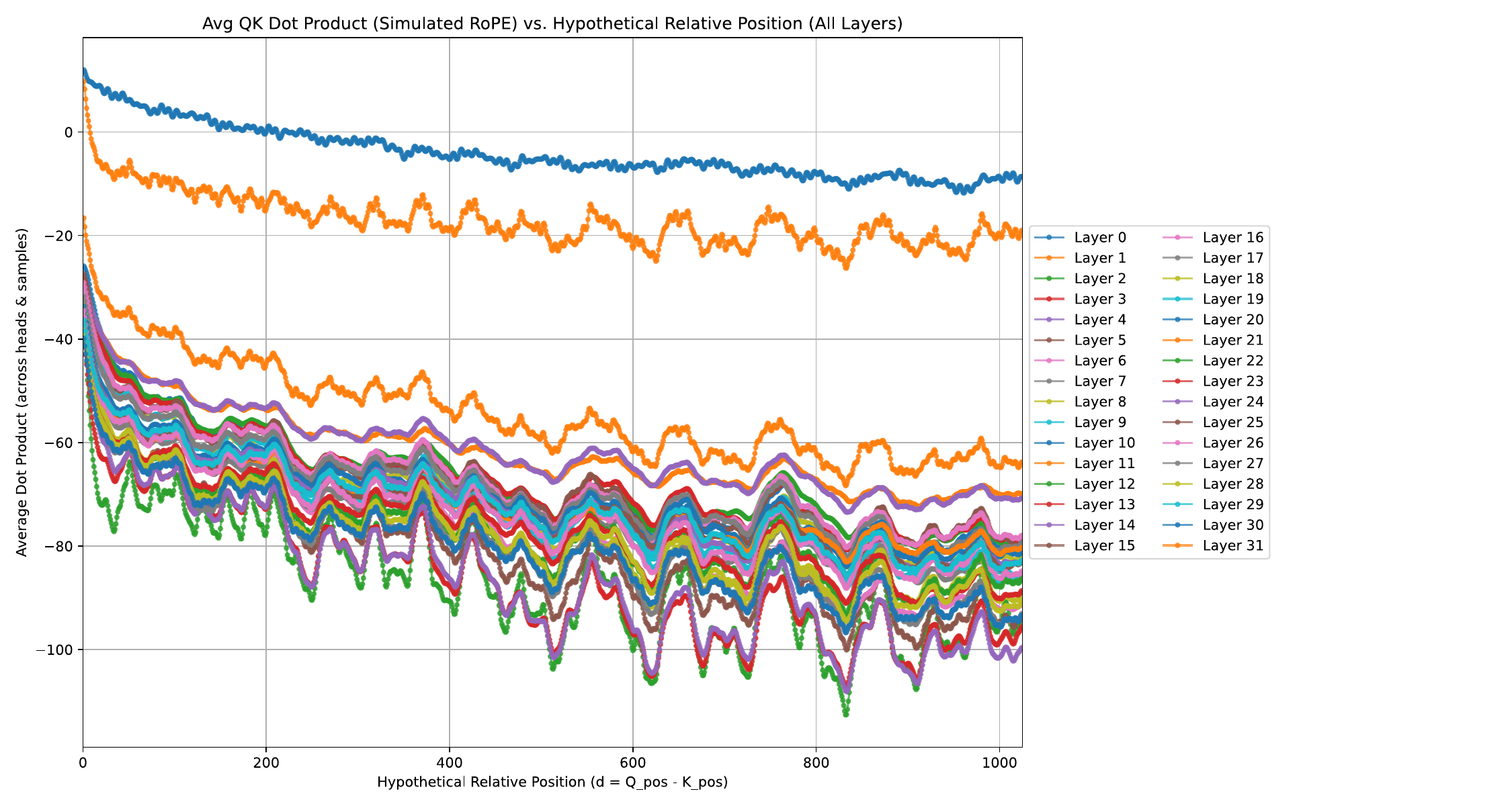}    
     \caption{The Long-term Decay Property of RoPE. We randomly sampled 100 text data points from Wikitext and randomly selected 10 pairs of q-k from each layer of the Vicuna-7B model for computation.}
     \label{fig:RoPE}
 \end{figure*}

\section{Method details}
\label{sec:method details}
Through the reorganization of position IDs, the "distance" between thumbnail tokens and their corresponding high-resolution tokens is reduced. This adjustment not only brings related embeddings closer in terms of positional encoding but also effectively restricts the growth of position IDs. Consequently, this approach prevents the issue of position IDs increasing by thousands when processing a single image, which could otherwise lead to exceeding the maximum position ID values encountered during training.

Our algorithm process is shown in Algorithm \ref{alg:multimodal_position}. In practice, assuming that the 2D feature map obtained after encoding the thumbnail with ViT has dimensions $(H_0,W_0)$, and the feature map obtained after encoding the entire high-resolution image has dimensions $(H_1,W_1)$, for simplicity, we assume that the positional ID of the first token in the thumbnail image is 0. We first generate a 1D tensor ranging from $0$ to $H_0 \times W_0$, then reshape it to $(H_0,W_0)$, and use interpolation to resize the reshaped tensor to 
$(H_1,W_1)$, rounding the values to integers. After flattening both parts, they are concatenated to form our positional IDs. 
\begin{algorithm}[ht]
\caption{ID-Align with RoPE}
\label{alg:multimodal_position}
\begin{algorithmic}[1]
\Require
\State $E_{\text{text}}$: Sequence of text embeddings
\State $E_{\text{low}}$: Sequence  of thumbnail embeddings
\State $E_{\text{high}}$: Sequence  of high-resolution image embeddings
\State $\mathcal{M}: E_{\text{high}} \to E_{\text{low}}$ : Return the  $E_{\text{low}}$ corresponding to $E_{\text{high}}$ 

\Ensure

\State $\textit{max\_pid} \gets 0$ 
\State $E_{\text{merged}} \gets \text{Concat}(E_{\text{text}}, E_{\text{low}}, E_{\text{high}})$

\For{each embedding $e_i \in E_{\text{merged}}$}
    \If{$e_i \in E_{\text{text}} \cup E_{\text{low}}$}
        \State $\text{pos\_id}(e_i) \gets \textit{max\_pid}$
        \State $\textit{max\_pid} \gets \textit{max\_pid} + 1$
    \ElsIf{$e_i \in E_{\text{high}}$}
        \State $\text{pos\_id}(e_i) \gets \text{pos\_id}(\mathcal{M}(e_i))$
        \State $\textit{max\_pid} \gets \max(\textit{max\_pid}, \mathcal{M}(e_i) + 1)$ 
    \EndIf
\EndFor

\Statex
\Function{ApplyRotaryEncoding}{$E_{\text{merged}}$}
\For{each $e_i \in E_{\text{merged}}$}
    \State $e_i \gets \text{RoPE}(e_i, \text{pos\_id}(e_i))$
\EndFor
\State \Return $E_{\text{merged}}$
\EndFunction
\end{algorithmic}
\end{algorithm}

\section{Experiments Setup}

\label{sec:experiments set up}

All experiments were conducted using eight A800 GPUs.
\subsection{Parameter Settings}

As for the hyperparameter settings, we adopted the configurations from Open-LLaVA-Next \citep{chen2024open}. We will also list these hyperparameters below.

\begin{lstlisting}[caption={The script for the LLaVA-Next pretrain phase, using Vicuna and CLIP as the LLM backbone and visual encoder, respectively.}]
nnodes=1
num_gpus=8
deepspeed --num_nodes ${nnodes} --num_gpus ${num_gpus} --master_port=10270 llava/train/train_mem.py \
    --deepspeed ./scripts/zero2.json \
    --model_name_or_path ${MODEL_PATH} \
    --version plain \
    --data_path ${DATA_PATH} \
    --image_folder ${IMAGE_FOLDER} \
    --vision_tower ${VISION_TOWER} \
    --mm_projector_type mlp2x_gelu \
    --tune_mm_mlp_adapter True \
    --unfreeze_mm_vision_tower False \
    --mm_vision_select_layer -2 \
    --mm_use_im_start_end False \
    --mm_use_im_patch_token False \
    --mm_patch_merge_type spatial_unpad \
    --image_aspect_ratio anyres \
    --group_by_modality_length False \
    --bf16 True \
    --output_dir ./checkpoints/${RUN_NAME} \
    --num_train_epochs 1 \
    --per_device_train_batch_size 8 \
    --per_device_eval_batch_size 4 \
    --gradient_accumulation_steps 4 \
    --evaluation_strategy "no" \
    --image_grid_pinpoints "[(336, 672), (672, 336), (672, 672), (1008, 336), (336, 1008)]" \
    --use_id_align True \
    --save_strategy "steps" \
    --save_steps 24000 \
    --save_total_limit 1 \
    --learning_rate 1e-3 \
    --weight_decay 0. \
    --warmup_ratio 0.03 \
    --lr_scheduler_type "cosine" \
    --logging_steps 1 \
    --tf32 True \
    --model_max_length 4096 \
    --gradient_checkpointing True \
    --dataloader_num_workers 4 \
    --lazy_preprocess True \
    --report_to None \
    --run_name ${RUN_NAME}
\end{lstlisting}

\begin{lstlisting}[caption={The script for the LLaVA-Next finetune phase, using Vicuna and CLIP as the LLM backbone and visual encoder, respectively.}]
nnodes=1
num_gpus=8

deepspeed --num_nodes ${nnodes} --num_gpus ${num_gpus} --master_port=10271 llava/train/train_mem.py \
    --deepspeed ./scripts/zero3.json \
    --model_name_or_path ${MODEL_PATH} \
    --version v1 \
    --data_path ${DATA_PATH} \
    --image_folder ${IMAGE_FOLDER} \
    --pretrain_mm_mlp_adapter ./checkpoints/${BASE_RUN_NAME}/mm_projector.bin \
    --unfreeze_mm_vision_tower True \
    --mm_vision_tower_lr 2e-6 \
    --vision_tower ${VISION_TOWER} \
    --mm_projector_type mlp2x_gelu \
    --mm_vision_select_layer -2 \
    --mm_use_im_start_end False \
    --use_id_align True \
    --mm_use_im_patch_token False \
    --group_by_modality_length True \
    --image_aspect_ratio anyres \
    --mm_patch_merge_type spatial_unpad \
    --bf16 True \
    --image_grid_pinpoints "[(336, 672), (672, 336), (672, 672), (1008, 336), (336, 1008)]" \
    --output_dir ./checkpoints/${RUN_NAME} \
    --num_train_epochs 1 \
    --per_device_train_batch_size 8 \
    --per_device_eval_batch_size 4 \
    --gradient_accumulation_steps 2 \
    --evaluation_strategy "no" \
    --save_strategy "steps" \
    --save_steps 1000 \
    --save_total_limit 1 \
    --learning_rate 2e-5 \
    --weight_decay 0. \
    --warmup_ratio 0.03 \
    --lr_scheduler_type "cosine" \
    --logging_steps 1 \
    --tf32 True \
    --model_max_length 4096 \
    --gradient_checkpointing True \
    --dataloader_num_workers 4 \
    --lazy_preprocess True \
    --report_to none \
    --run_name ${RUN_NAME}
\end{lstlisting}
\begin{lstlisting}[caption={The script for the LLaVA-Next pre-train phase, using Qwen and SigLIP as the LLM backbone and visual encoder, respectively.}]
nnodes=1
num_gpus=8
deepspeed --num_nodes ${nnodes} --num_gpus ${num_gpus} --master_port=10270 llava/train/train_mem.py \
    --deepspeed ./scripts/zero2.json \
    --model_name_or_path ${MODEL_PATH} \
    --version plain \
    --data_path ${DATA_PATH} \
    --image_folder ${IMAGE_FOLDER} \
    --vision_tower ${VISION_TOWER} \
    --mm_projector_type mlp2x_gelu \
    --tune_mm_mlp_adapter True \
    --unfreeze_mm_vision_tower False \
    --mm_vision_select_layer -2 \
    --mm_use_im_start_end False \
    --mm_use_im_patch_token False \
    --mm_patch_merge_type spatial_unpad \
    --image_aspect_ratio anyres \
    --group_by_modality_length False \
    --bf16 True \
    --output_dir ./checkpoints/${RUN_NAME} \
    --num_train_epochs 1 \
    --per_device_train_batch_size 8 \
    --per_device_eval_batch_size 4 \
    --gradient_accumulation_steps 4 \
    --evaluation_strategy "no" \
    --image_grid_pinpoints "[(384, 768), (768, 384), (768, 768), (1152, 384), (384, 1152)]" \
    --use_id_align True \
    --save_strategy "steps" \
    --save_steps 24000 \
    --save_total_limit 1 \
    --learning_rate 1e-3 \
    --weight_decay 0. \
    --warmup_ratio 0.03 \
    --lr_scheduler_type "cosine" \
    --logging_steps 1 \
    --tf32 True \
    --model_max_length 32768 \
    --gradient_checkpointing True \
    --dataloader_num_workers 4 \
    --lazy_preprocess True \
    --report_to none \
    --run_name ${RUN_NAME}
\end{lstlisting}

\begin{lstlisting}[caption={The script for the LLaVA-Next finetune phase, using Qwen and SigLIP as the LLM backbone and visual encoder, respectively}]
nnodes=1
num_gpus=8
deepspeed --num_nodes ${nnodes} --num_gpus ${num_gpus} --master_port=10271 llava/train/train_mem.py \
    --deepspeed ./scripts/zero3.json \
    --model_name_or_path ${MODEL_PATH} \
    --version ${PROMPT_VERSION} \
    --data_path ${DATA_PATH} \
    --image_folder ${IMAGE_FOLDER} \
    --pretrain_mm_mlp_adapter ./checkpoints/${BASE_RUN_NAME}/mm_projector.bin \
    --unfreeze_mm_vision_tower True \
    --mm_vision_tower_lr 2e-6 \
    --vision_tower ${VISION_TOWER} \
    --mm_projector_type mlp2x_gelu \
    --mm_vision_select_layer -2 \
    --mm_use_im_start_end False \
    --use_id_align True \
    --mm_use_im_patch_token False \
    --group_by_modality_length True \
    --image_aspect_ratio anyres \
    --mm_patch_merge_type spatial_unpad \
    --bf16 True \
    --image_grid_pinpoints "[(384, 768), (768, 384), (768, 768), (1152, 384), (384, 1152)]" \
    --output_dir ./checkpoints/${RUN_NAME} \
    --num_train_epochs 1 \
    --per_device_train_batch_size 8 \
    --per_device_eval_batch_size 4 \
    --gradient_accumulation_steps 2 \
    --evaluation_strategy "no" \
    --save_strategy "steps" \
    --save_steps 1000 \
    --save_total_limit 1 \
    --learning_rate 2e-5 \
    --weight_decay 0. \
    --warmup_ratio 0.03 \
    --lr_scheduler_type "cosine" \
    --logging_steps 1 \
    --tf32 True \
    --model_max_length 32768 \
    --gradient_checkpointing True \
    --dataloader_num_workers 4 \
    --lazy_preprocess True \
    --report_to none \
    --run_name ${RUN_NAME}
\end{lstlisting}

\subsection{Benchmarks}
Focusing on the overall and various hierarchical capabilities of models, we primarily adopted three benchmarks—MMBench \citep{liu2024mmbench}, MME \citep{yin2023survey}, and MMStar \citep{chen2024we}. Additionally, SeedBench-2-Plus \citep{li2024seed} and AI2D \citep{kembhavi2016diagram} were utilized to assess the models' capability in processing rich text images such as charts, maps, and web pages. RealWorldQA was employed to evaluate the models' effectiveness in handling real-world images, whereas POPE \citep{li2023evaluating} was used to examine the phenomenon of model hallucinations. To evaluate the model's performance on QA tasks, we will utilize the VQAv2 \citep{goyal2017making} and ScienceQA \citep{lu2022learn} datasets. We utilized LMMS-Eval \citep{zhang2024lmmsevalrealitycheckevaluation} for the evaluation of our model.
The decision to utilize ID-Align can be controlled by setting the value of \texttt{use\_id\_align} in the training scripts.

\section{More Results and Analysis}

\subsection{Learning Curve}

 In this section, we also plot the learning curve. From these curves, it can be observed that after applying ID-Align, the training loss is slightly lower during the latter half of the training phase compared to when not using ID-Align. Additionally, the gradient norm is notably lower, indicating that the model is closer to achieving convergence. This effect is especially pronounced on Vicuna. These plots were generated using a sliding average window with a window length of 100.
\label{sec: Learning Curve}
\subsubsection{Vicuna}
\begin{figure}[H]
    \centering
    \includegraphics[width=1\linewidth]{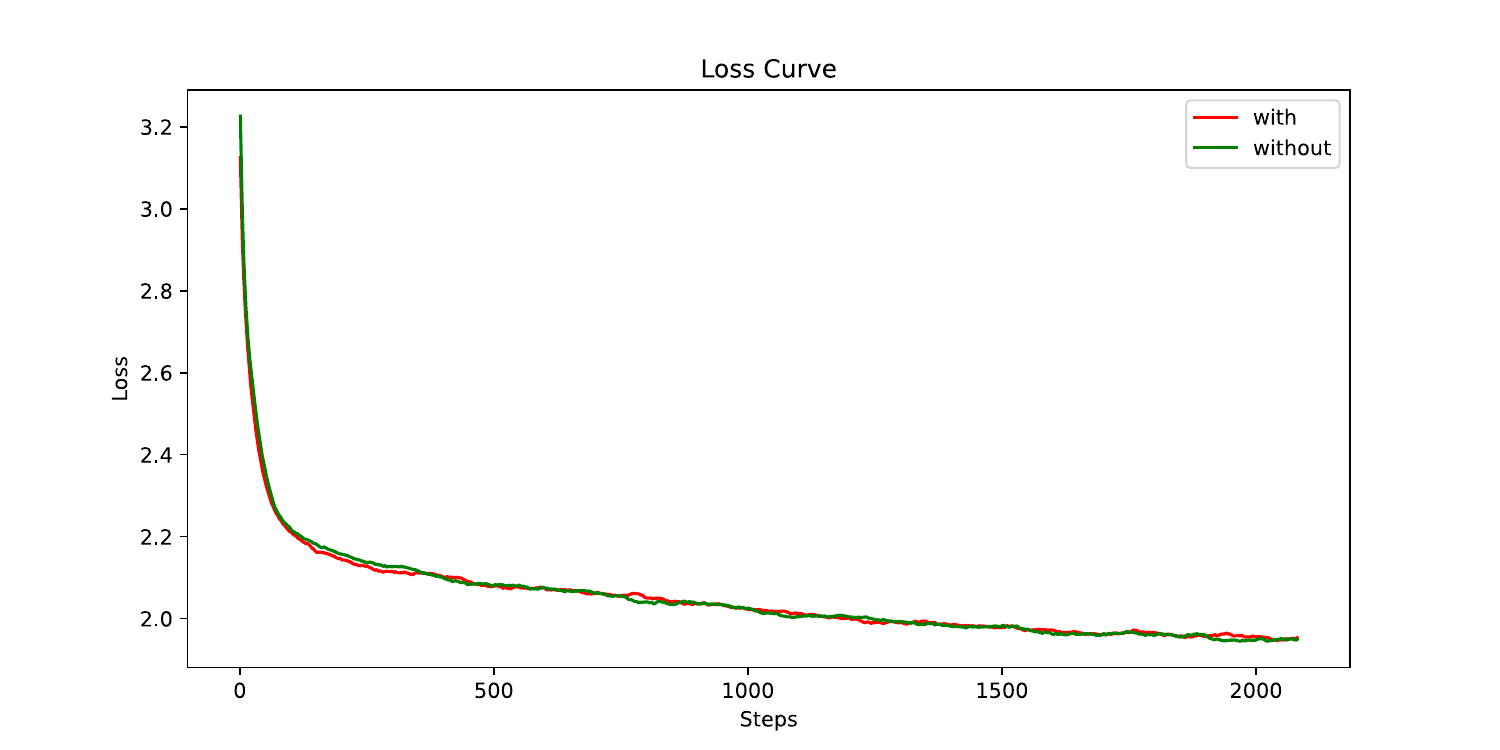}
    \caption{Pretrain Loss}
    \label{fig:vicuna-pretrain-loss}
\end{figure}
\begin{figure}[H]
    \centering
    \includegraphics[width=1\linewidth]{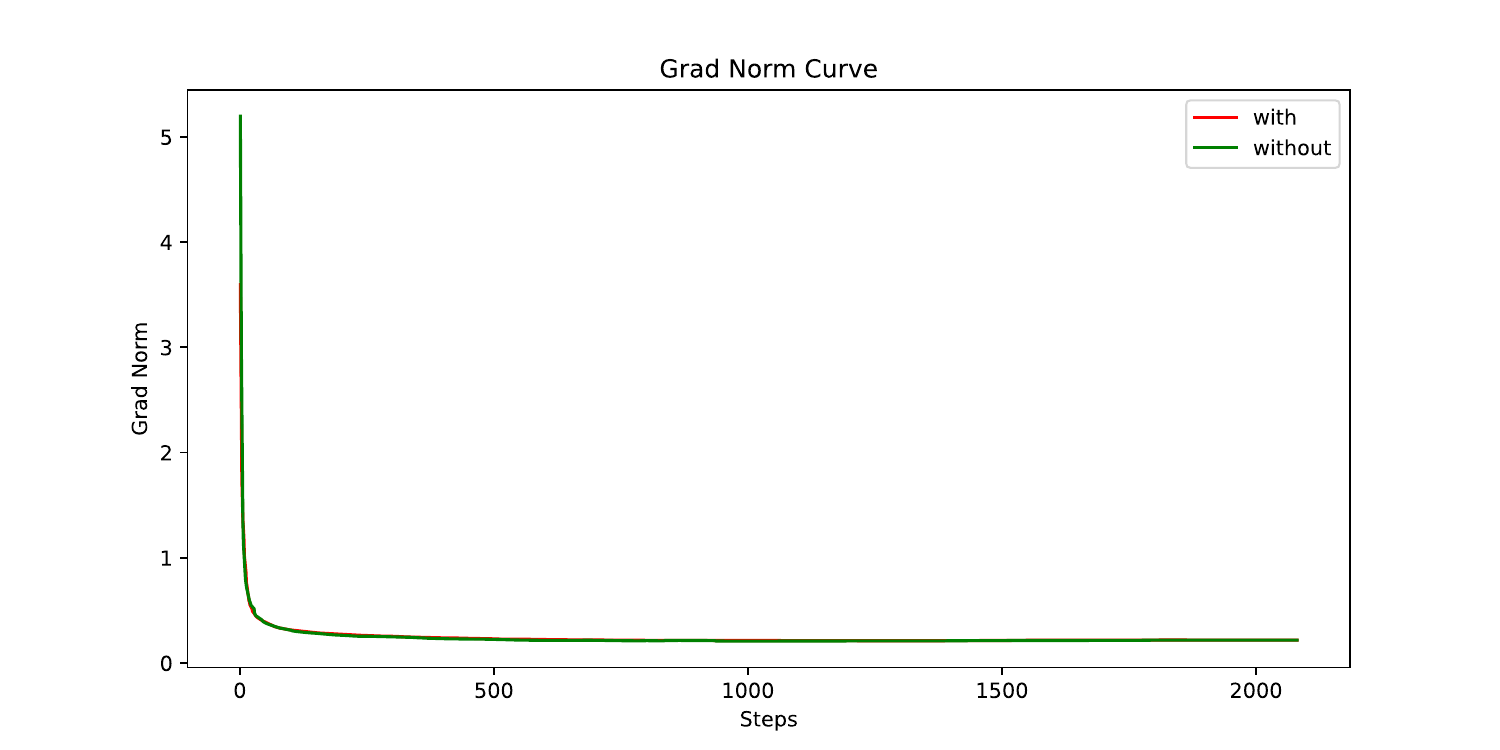}
    \caption{Pretrain Grad Norm}
    \label{fig:vicuna-pretrain-grad}
\end{figure}
\begin{figure}[H]
    \centering
    \includegraphics[width=1\linewidth]{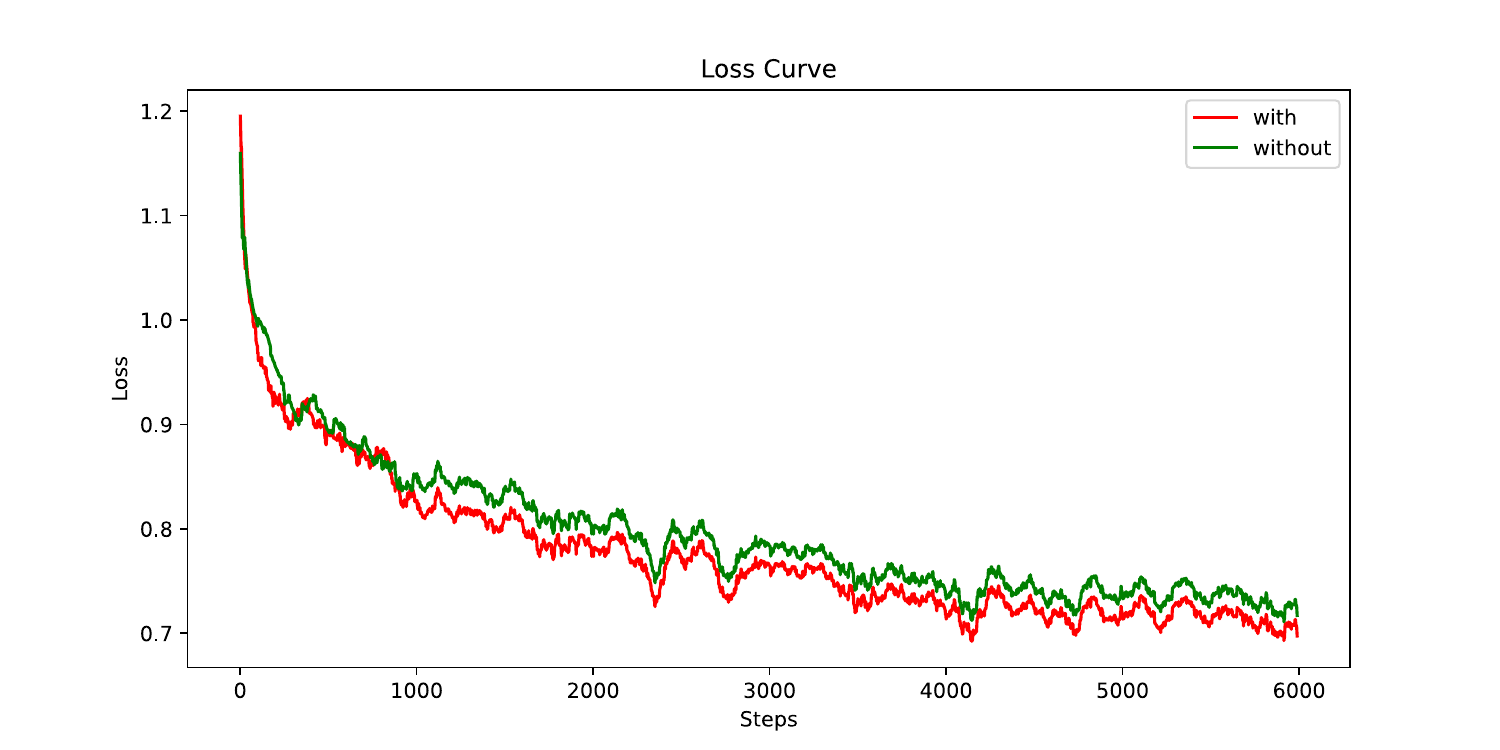}
    \caption{Finetune Loss}
    \label{fig:vicuna-finetune-loss}
\end{figure}
\begin{figure}[H]
    \centering
    \includegraphics[width=1\linewidth]{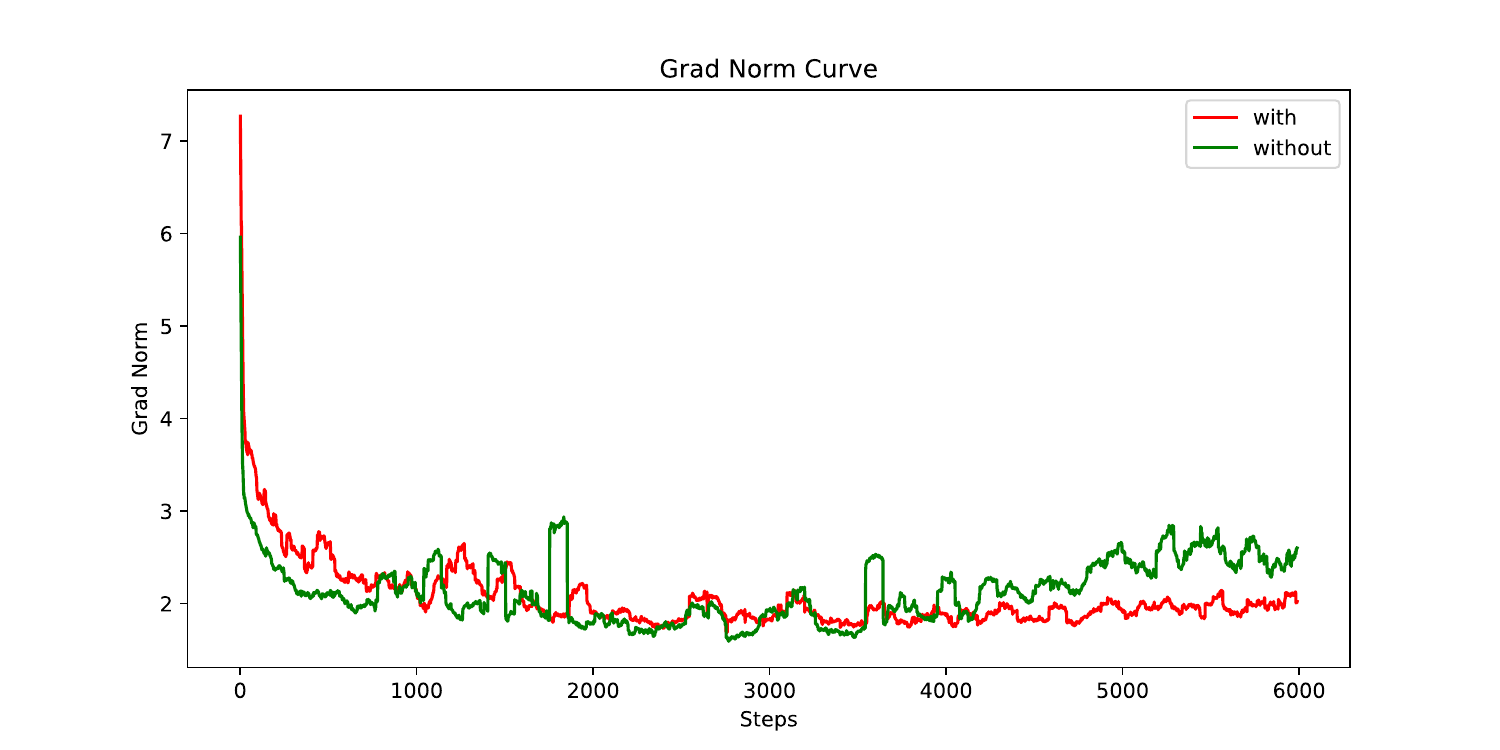}
    \caption{Finetune Grad Norm}
    \label{fig:vicuna-finetune-grad}
\end{figure}
\subsubsection{Qwen}
\begin{figure}[H]
    \centering
    \includegraphics[width=1\linewidth]{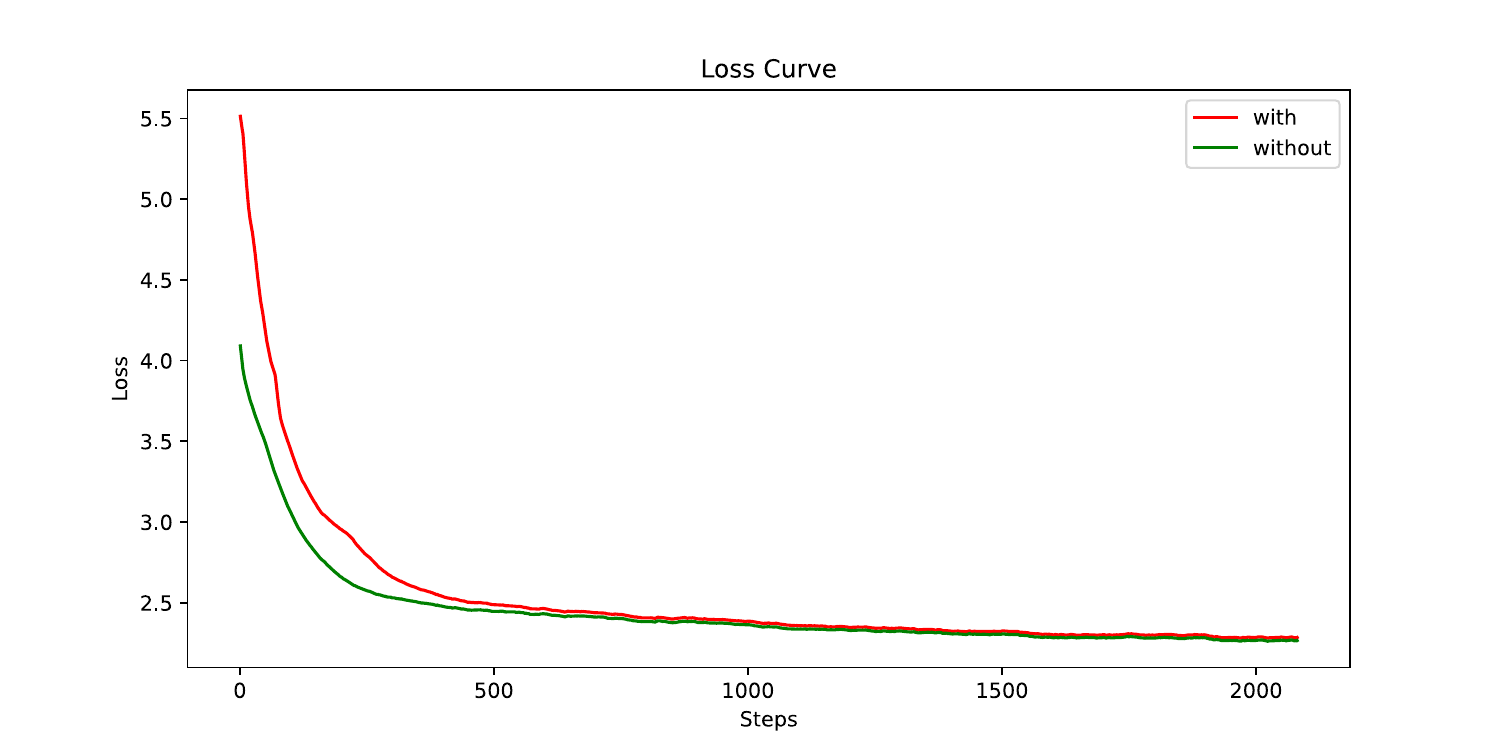}
    \caption{Pretrain Loss}
    \label{fig:qwen-pretrain-loss}
\end{figure}
\begin{figure}[H]
    \centering
    \includegraphics[width=1\linewidth]{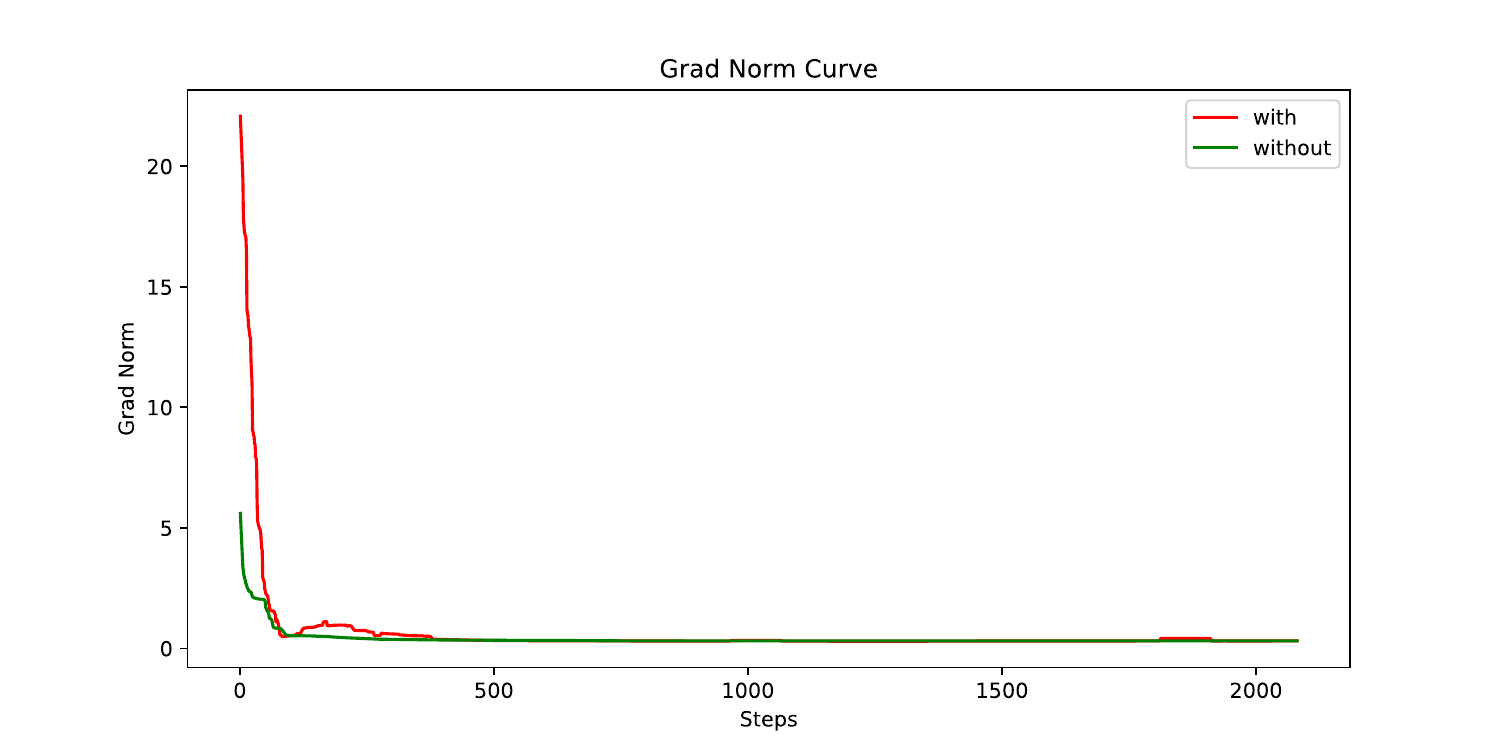}
    \caption{Pretrain Grad Norm}
    \label{fig:qwen-pretrain-grad}
\end{figure}
\begin{figure}[H]
    \centering
    \includegraphics[width=1\linewidth]{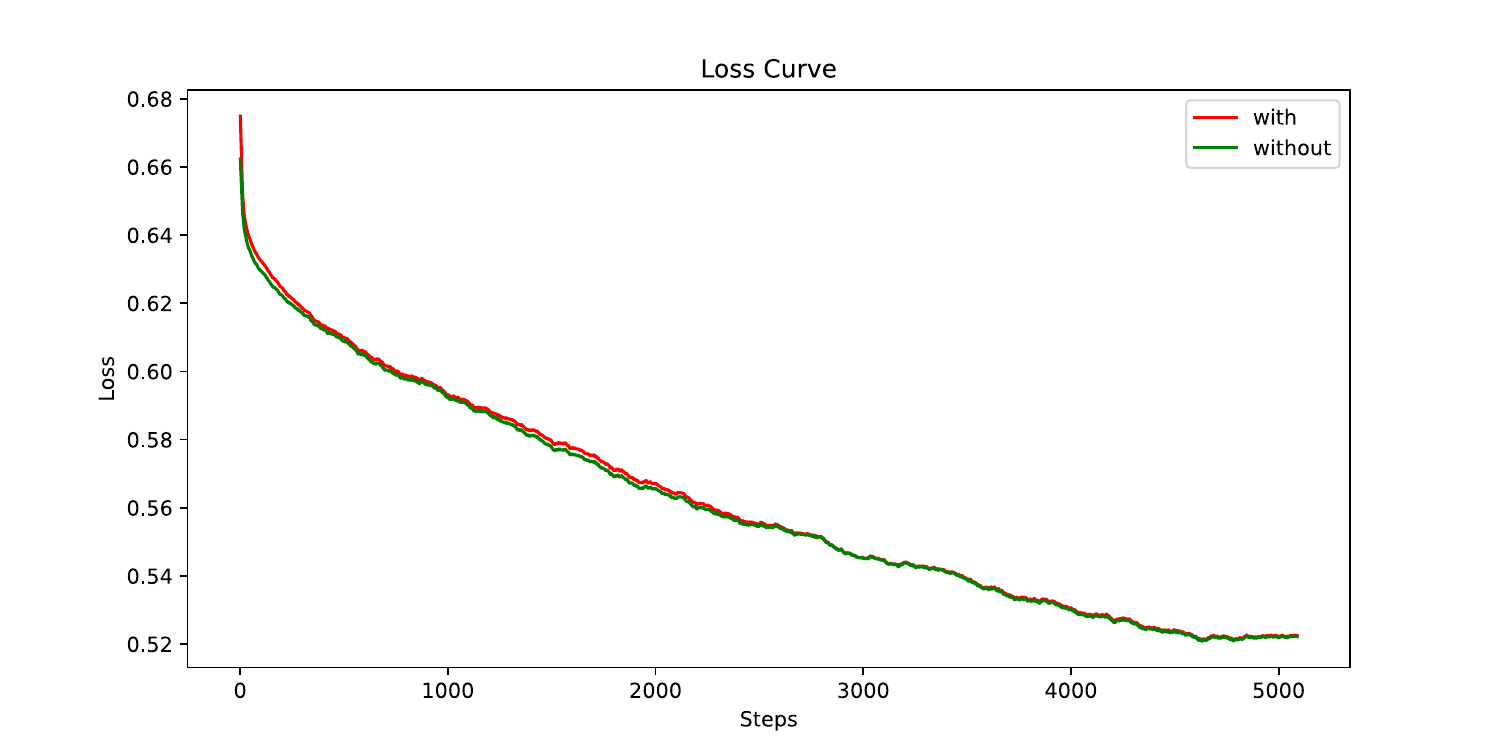}
    \caption{Finetune Loss}
    \label{fig:qwen-finetune-loss}
\end{figure}
\begin{figure}[H]
    \centering
    \includegraphics[width=1\linewidth]{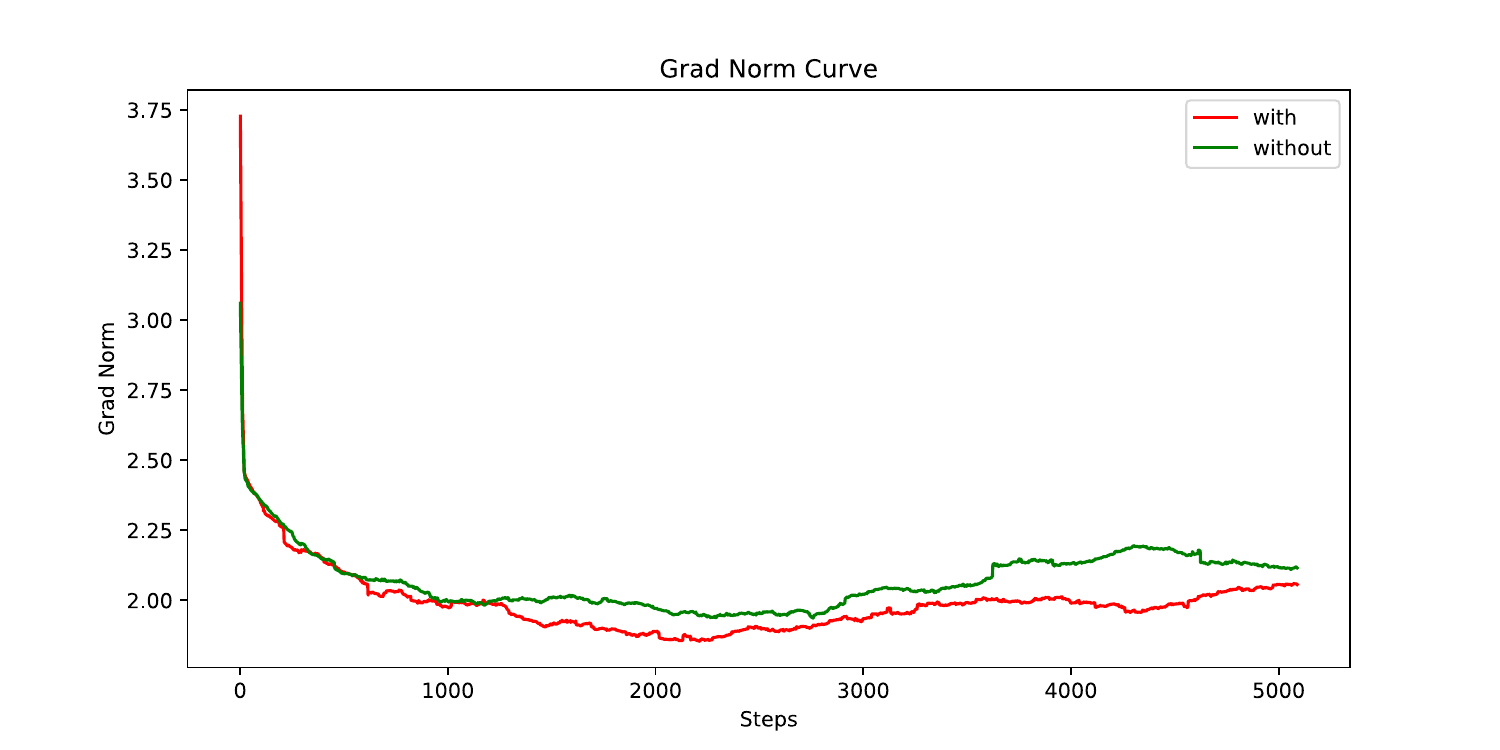}
    \caption{Finetune Grad Norm}
    \label{fig:qwen-finetune-grad}
\end{figure}
\subsection{Compare with Other Methods}

We also compared our method with MRoPE \citep{wang2024qwen2} and V2PE \citep{Ge2024V2PEIM}. Our method is not in competition with these methods; rather, it is compatible with them. The focus of these methods is on positional encodings within a single image, whereas our method addresses the correspondence between thumbnail and high-resolution images. Therefore, these methods can be combined.  

Due to the limitation of computing resource, we only experimented with the Qwen-2.5-0.5B model and SigLip. These experiments only involved adjusting these two methods to suit high-resolution scenarios, without combining them with ID-Align. For V2PE, we set $\epsilon=0.5$, which is the best result reported in their paper for conventional benchmarks. For MRoPE, we treated the thumbnail and high-resolution images as separate images. Since the hidden state dimension of the 0.5B model is small, we set the MRoPE section to $[8,12,12]$, which is a proportionally scaled version of the $[16,24,24]$ used in the Qwen2.5-VL-3B model. The results are shown in Table \ref{tab:v2pe}.

Compared to V2PE and MRoPE, our method shows significant improvement in metrics that measure the overall capability of the model (MME, MMBench, MMStar). In metrics that measure specific capabilities, such as PoPE and AI2D, our method may not perform as well as V2PE or MRoPE, which could be related to the characteristics of their methods and the benchmark data distribution. Overall, in the context of dynamic high-resolution, our method is superior. 

\begin{table*}[!htbp]
\centering
\small
\setlength{\tabcolsep}{3.5pt}
\begin{tabular}{lccccccccccc}
\toprule
       & MMB   & MMStar & RWQA  & SEEDB & POPE  & MME-C & MME-P & AI2D  & VQAV2 & SQA   & avg   \\
\midrule
V2PE   & 56.28 & 37.43  & 51.76 & \textbf{48.09} & 87.82 & 30.85 & 63.86 & \textbf{57.87} & \textbf{65.62} & 60.83 & 56.04 \\
MRoPE  & 55.44 & 38.28  & 53.73 & 46.73 & \textbf{88.41} & 30.01 & 62.81 & 57.77 & 64.78 & 59.89 & 55.79 \\
ID-Align & \textbf{57.68} & \textbf{39.74} & \textbf{55.03} & 47.56 & 87.50 & \textbf{31.03} & \textbf{64.03} & 56.96 & 64.86 & \textbf{60.88} & \textbf{56.53} \\
\bottomrule
\end{tabular}
\caption{Comparison with MRoPE and V2PE} % Add a caption
\label{tab:v2pe} % Add a label for referencing
\end{table*}
\subsection{MMBench Leaf Tasks}
\label{sec:mmbench}
\textbf{Coarse Perception:}
\begin{itemize}
    \item Image Style
    \item Image Topic
    \item Image Scene
    \item Image Mood
    \item Image Quality
\end{itemize}

\textbf{Fine-grained Perception (Single-instance):}
\begin{itemize}
    \item Attribute Recognition
            \item Celebrity Recognition
    \item Object Localization
    \item OCR
\end{itemize}

\textbf{Fine-grained Perception (Cross-instance):}
\begin{itemize}
    \item Spatial Relationship
            \item Attribute Comparison
            \item Action Recognition
\end{itemize}

\textbf{Attribute Reasoning:}
\begin{itemize}
    \item Physical Property Reasoning
            \item Function Reasoning
            \item Identity Reasoning
\end{itemize}

\textbf{Relation Reasoning:}
\begin{itemize}
    \item Social Relation
    \item Nature Relation
            \item Physical Relation
\end{itemize}

\textbf{Logic Reasoning:}
\begin{itemize}
    \item Future Prediction
    \item Structuralized Image-text Understanding
\end{itemize}

\end{document}